\documentclass[11pt]{article}

\usepackage[preprint]{acl}

\usepackage{times}
\usepackage{latexsym}

\usepackage[T1]{fontenc}
\usepackage[utf8]{inputenc}

\usepackage{microtype}

\usepackage{inconsolata}

\usepackage[T1]{fontenc}
\usepackage[utf8]{inputenc}
\usepackage{CJKutf8}
\usepackage{booktabs}

\usepackage{url}
\usepackage{graphicx}
\usepackage{float}
\usepackage{flafter}
\usepackage{wrapfig}
\usepackage{microtype}

\usepackage[table]{xcolor}

\usepackage{booktabs}
\usepackage{array}
\usepackage{tabularx}
\usepackage{multirow}
\usepackage{multicol}
\usepackage{makecell}
\usepackage{colortbl}
\usepackage{arydshln}
\usepackage{needspace}
\expandafter\let\csname endtabular*\endcsname\endtabular
\usepackage{tikz}
\usepackage[most]{tcolorbox}
\usepackage{listings}
\usepackage{subcaption}

\usepackage{algorithm}
\usepackage{algpseudocode}

\usepackage{hyperref}
\usepackage{amsmath,amsfonts,bm}

\def\eqref#1{equation~\ref{#1}}
\def\1{\bm{1}}

\DeclareMathAlphabet{\mathsfit}{\encodingdefault}{\sfdefault}{m}{sl}
\SetMathAlphabet{\mathsfit}{bold}{\encodingdefault}{\sfdefault}{bx}{n}

\usepackage[most]{tcolorbox}
\usepackage{listings}
\usepackage{xcolor}
\usepackage{enumitem}

\definecolor{promptblack}{RGB}{0,0,0}
\definecolor{promptbody}{RGB}{244,244,244}

\lstdefinestyle{promptstyle}{%
  basicstyle=\ttfamily\footnotesize,
  columns=fullflexible,
  keepspaces=true,
  breaklines=true,
  breakatwhitespace=false,
  showstringspaces=false,
  frame=none,
  lineskip=1pt,
  aboveskip=0pt,
  belowskip=0pt,
  inputencoding=utf8,
  extendedchars=true,
}

\tcbset{promptboxbase/.style={%
  enhanced jigsaw,
  breakable,
  colback=promptbody,
  colframe=promptblack,
  boxrule=1.25pt,
  arc=5pt,
  outer arc=5pt,
  toprule at break=1.25pt,
  bottomrule at break=1.25pt,
  topsep at break=0pt,
  bottomsep at break=0pt,
  overlay first and middle={
    \draw[promptblack,line width=1.25pt,rounded corners=5pt]
      (frame.north west) rectangle (frame.south east);
  },
  overlay middle and last={
    \draw[promptblack,line width=1.25pt,rounded corners=5pt]
      (frame.north west) rectangle (frame.south east);
  },
  left=12pt,
  right=12pt,
  top=10pt,
  bottom=10pt,
  colbacktitle=promptblack,
  coltitle=white,
  fonttitle=\rmfamily\bfseries\small,
  title filled,
  boxed title style={
    boxrule=0pt,
    sharp corners,
    colback=promptblack,
    left=12pt,
    right=12pt,
    top=5pt,
    bottom=5pt,
  },
}}

\newtcblisting{promptbox}[1][PROMPT / MARKDOWN]{%
  promptboxbase,
  title={\detokenize{#1}},
  listing only,
  listing options={style=promptstyle},
}

\newtcolorbox{prompttext}[1][PROMPT]{%
  promptboxbase,
  title={\detokenize{#1}},
}

\input{case_study}

\definecolor{rowgray}{gray}{0.95}

\definecolor{traj}{HTML}{4E79A7}
\definecolor{prog}{HTML}{59A14F}
\definecolor{buck}{HTML}{C45A4A}

\definecolor{trajbg}{HTML}{DCE6F2}
\definecolor{progbg}{HTML}{DCEFEA}
\definecolor{buckbg}{HTML}{F3DFDA}

\definecolor{orangebg}{HTML}{F4E6D5}

\definecolor{purplebg}{HTML}{E8E0F0}

\definecolor{tabPurple}{RGB}{151,136,177}
\definecolor{tabBlue}{RGB}{113,145,178}
\definecolor{tabGreen}{RGB}{126,159,137}
\definecolor{tabOrange}{RGB}{201,158,112}
\definecolor{tabMagenta}{RGB}{184,132,158}

\newlength{\metricbarw}
\newcommand{\metricbar}[3]{%
  \makebox[0pt][l]{%
    \textcolor{#2!32}{%
      \rule[-0.55ex]{#1\metricbarw}{2.25ex}%
    }%
  }%
  \makebox[\metricbarw][r]{#3\,}%
}

\newcommand{\iterkey}[2]{%
  \raisebox{0.12ex}{%
    \textcolor{#1!85}{%
      \rule{0.68ex}{0.68ex}%
    }%
  }%
  \hspace{0.45em}#2%
}

\definecolor{EverModelBlue}{RGB}{235,242,250}

\providecommand{\evergptmodelicon}{}
\providecommand{\everdeepseekmodelicon}{}
\providecommand{\evermodelbarstrut}{}
\providecommand{\evergroupspacedash}{}

\renewcommand{\evermodelbarstrut}{%
  \rule[-0.72ex]{0pt}{2.40ex}%
}

\renewcommand{\evergroupspacedash}{%
  \hdashline
}

\definecolor{EvoBlue}{HTML}{4E79A7}
\definecolor{EvoGreen}{HTML}{59A14F}
\definecolor{EvoRed}{HTML}{C65D4B}
\definecolor{EvoOrange}{HTML}{E39C45}
\definecolor{EvoPurple}{HTML}{9C6ADE}

\newlength{\EvoMetricBarWidth}
\newlength{\EvoMetricFillWidth}
\newlength{\EvoMetricTextWidth}

\newcommand{\EvoMetricBar}[3]{%
  \settowidth{\EvoMetricTextWidth}{%
    {#3}\hspace{0.34em}%
  }%
  \setlength{\EvoMetricFillWidth}{#2\EvoMetricBarWidth}%
  \ifdim\EvoMetricFillWidth<\EvoMetricTextWidth
    \setlength{\EvoMetricFillWidth}{\EvoMetricTextWidth}%
  \fi
  \makebox[0pt][l]{%
    \textcolor{#1!24}{%
      \rule[-0.58ex]{\EvoMetricFillWidth}{2.30ex}%
    }%
  }%
  \makebox[\EvoMetricBarWidth][r]{%
    {#3}\hspace{0.22em}%
  }%
}

\newcommand{\EvoIterationKey}[2]{%
  \raisebox{0.42ex}{%
    \textcolor{#1}{%
      \rule{1.08em}{1.55pt}%
    }%
  }%
  \hspace{0.48em}#2%
}

\definecolor{GroupBlue}{RGB}{225,235,245}
\definecolor{GroupPeach}{RGB}{250,232,220}

\definecolor{AblBlue}{HTML}{3B70B9}
\definecolor{AblTeal}{HTML}{269B91}
\definecolor{AblRed}{HTML}{C85845}

\newlength{\ablscorew}
\newlength{\abldeltaw}
\usepackage{graphicx}
\usepackage{booktabs}
\usepackage{array}
\usepackage{caption}

\newcommand{\EvoCellBar}[3]{%
  \begingroup
  \leavevmode
  \makebox[\hsize][c]{%
    \rlap{%
      \textcolor{#1!22}{%
        \rule{#2\hsize}{1.45ex}%
      }%
    }%
    \makebox[\hsize][c]{\textcolor{black}{#3}}%
  }%
  \endgroup
}

\usepackage{placeins}

\title{EvoGroup: A Self-Evolving Harness for Group Agents in Shared Contexts}

\author{First Author \\
  Affiliation / Address line 1 \\
  Affiliation / Address line 2 \\
  Affiliation / Address line 3 \\
  \texttt{email@domain} \\\And
  Second Author \\
  Affiliation / Address line 1 \\
  Affiliation / Address line 2 \\
  Affiliation / Address line 3 \\
  \texttt{email@domain} \\}

\author{
 \textbf{Xinyi Xu\textsuperscript{1}},
 \textbf{Xiaoyang Li\textsuperscript{1}},
 \textbf{Yucheng Feng\textsuperscript{1}},
 \textbf{Wenqing Chen\textsuperscript{1}\thanks{Corresponding author.}},
 \textbf{Youliang Yuan\textsuperscript{2}},
\\
 \textbf{Wei Shi\textsuperscript{2}},
 \textbf{Rongxiang Weng\textsuperscript{2}},
 \textbf{Zibin Zheng\textsuperscript{1}}
\\[0.5em]
 \textsuperscript{1}Sun Yat-sen University\\
 \textsuperscript{2}Longcat Team, Meituan Inc.
\\
 \texttt{xuxy299@mail2.sysu.edu.cn}
}
\begin{document}
\maketitle
\begin{abstract}
With the rise of Group Agent such as Claude Tag, agents increasingly serve as digital workers for long-term team collaboration. 
These agents operate over shared contexts that continuously accumulate through multi-user interactions. 
However, existing agent memory systems are mainly designed for individual users and struggle to capture group-level knowledge in shared contexts, including group relationships and collaboratively formed knowledge.
Manual memory strategies further limit adaptation to diverse and evolving shared contexts.
To this end, we propose \textbf{EvoGroup (EvoG)}, a self-evolving harness for group agents that starts from a minimal six-tool cold start and adapts through interaction with shared context.
EvoG uses confidence and task outcomes to select informative trajectories, while also eliciting self-reflections on uncertainty in low-confidence cases.
The analysis agent then examines selected trajectories through adaptive progressive disclosure, followed by bucketed analysis to synthesize high-level findings that guide the evolve agent in revising the Harness.
Empirically, pass@1 on held-out split of EverMemBench improves from 71.07\% to 89.58\%, surpassing both self-evolving memory and human-designed harnesses, including Claude Code (85.83\%). 
To evaluate generalization, we freeze the evolved harness and evaluate it on the unseen GroupMemBench, where, with GPT-5.5, it demonstrates effective generalization and outperforms both memory baselines and human-designed harnesses.\footnote{The code for this manuscript is available at 
\href{https://github.com/BigggFatttRabittt/EvoGroup}{https://github.com/BigggFatttRabittt/EvoGroup}.}
\end{abstract}

\begin{figure}[t]
\centering
\includegraphics[width=\linewidth]{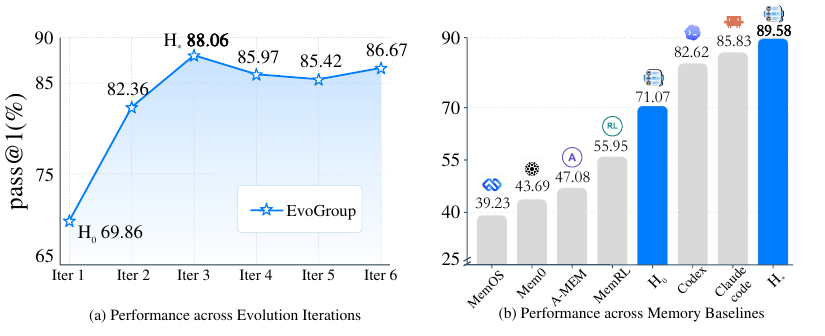}
\caption{\textbf{Performance of EvoGroup with DeepSeek-V4-Flash on EverMemBench}. (a) Evolution over 5 iterations on the held-in split, improving pass@1 by $18.2$ pp from $H_0$ to $H_*$. (b) Comparison on the held-out split, where $H_*$ outperforms all the baselines including Codex and Claude Code.}
\label{fig:figure1}
\vspace{-1em}
\end{figure}

\section{Introduction}

AI agents are gradually evolving from conversational assistants designed for individual users into digital workers that directly participate in multi-user interactions and team collaboration \citep{openai2026workspaceagents,spataro2025humanagent,claudetag}. 
In sustained multi-user collaboration, shared contexts continuously accumulate and evolve through interactions among multiple users. 
This makes memory essential for agents to accurately extract and apply relevant information from long-term conversations. 
A Group Agent must not only understand current messages but also integrate discussion topics, key events, member roles, and group decisions from prior interactions. 

However, most existing agent memory systems are designed for single-user settings, organizing memory around individual experiences, preferences, and interaction histories \citep{park2023generativeagents,zhong2024memorybank,packer2023memgpt}. In group settings, information is distributed across multiple users, topics, and reply threads, and events may be supplemented or reinterpreted by different members. These characteristics make it difficult for existing memory systems to capture group dynamics and collaboratively formed knowledge \citep{yang2026groupmembenchbenchmarkingllmagent}.
In addition, existing memory systems typically predefine how memory is represented, updated, and retrieved \citep{chhikara2025mem0,liu2026simplemem}. 
For example, interaction histories may be stored in vector databases \citep{chhikara2025mem0} or organized as knowledge graphs \citep{anokhin2024arigraph}, with fixed mechanisms for updating outdated or conflicting information \citep{fang2026lightmem}. 
When group structures, shared contexts, or task requirements change, adapting these systems often requires redesigning their memory storage or retrieval mechanisms. Such predefined designs limit the ability to adapt to different groups.

To address these challenges, we propose \textbf{EvoGroup (EvoG)}, a self-evolving harness framework for Group Agent memory. 
\textbf{EvoG} starts from a minimal harness with only six tools and evolves through interaction.
The evolution loop consists of four stages. 
\textbf{(1) Interaction.} 
The Group Agent uses the harness to answer queries over accumulated shared contexts, producing trajectories and confidence scores. 
For low-confidence queries, the agent additionally reflects on the reasons for its uncertainty before accessing the ground-truth.
\textbf{(2) Trajectory Selection.} 
\textbf{EvoG} combines confidence with official evaluation to identify informative trajectories, including confident-incorrect, unconfident-correct, and unconfident-incorrect. 
\textbf{(3) Analysis.} 
The analysis agent uses Adaptive Progressive Disclosure (APD) to analyze selected trajectories, accessing trajectory details only when necessary. The agent also compresses detailed rationales into concise rationales.
To identify recurring issues across trajectories, the agent performs bucketed analysis by grouping rationales according to inferred query semantics and further aggregating them into high-level findings. \textbf{(4) Revision.} 
Finally, the evolve agent uses analysis outputs to revise the Harness, producing $H_{t+1}$ for the next round of interaction.
Through this loop, \textbf{EvoG} enables the harness to continuously adapt to the group and shared context.

\begin{figure*}[t]
\centering
\includegraphics[width=\linewidth]{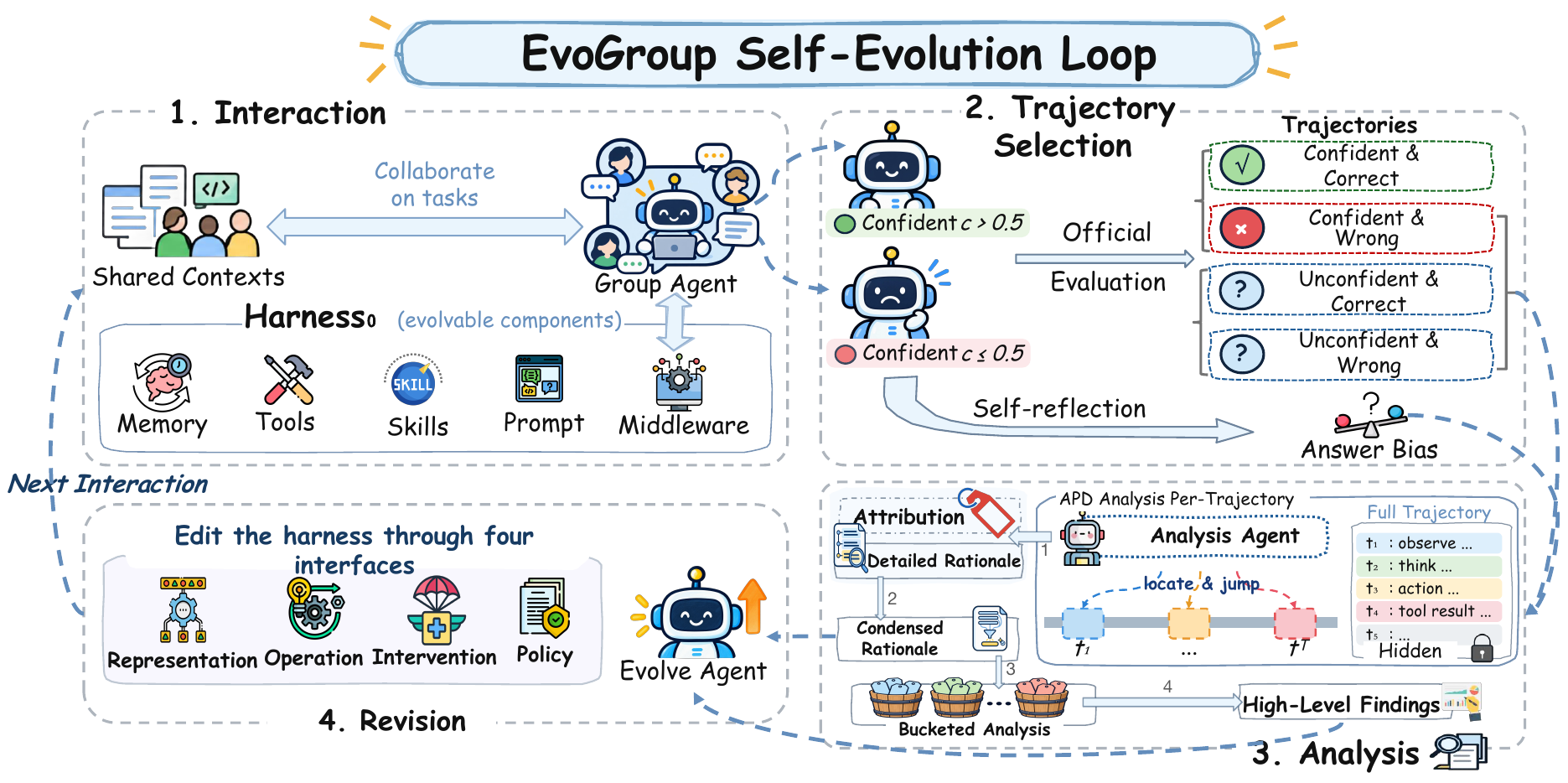}
\caption{\textbf{Overview of the EvoGroup loop}, which operates in four phases: I) Interaction, which Group Agent collaborates with shared contexts; \textbf{II) Trajectory Selection}, which thresholds trajectories by confidence; III) Analysis, which performs per-trajectory and bucketed analysis; and IV) Revision, which updates harness through four interfaces. }
\label{fig:figure2}
\end{figure*}

We conduct extensive experiments to evaluate the effectiveness of \textbf{EvoG}. 
After evolution, the pass@1 of DeepSeek-V4-Flash on held-out split of EverMemBench improves from 71.07\% to 89.58\%, outperforming the human-designed Harnesses Codex (82.62\%) and Claude Code (85.83\%), as well as the strongest self-evolving memory baseline (55.95\%). 
Across two additional backbone models, \textbf{EvoG} consistently improves pass@1 by 15.36--16.19 pp. 
To evaluate generalization beyond the evolution distribution, we freeze the evolved harness and apply it to GroupMemBench without further adaptation, where it continues to outperform self-evolving memory baselines and Claude Code. 
We further conduct component and process-level ablations and analyze capability transitions, confidence, self-reflection, APD analysis, and harness revisions to better understand how \textbf{EvoG} evolves.
Together with our findings and open-source release, we hope \textbf{EvoG} inspires future efforts toward more capable and adaptive Group Agents.

\section{EvoGroup}
In this section, we propose \textbf{EvoGroup (EvoG)}, a self-evolving harness loop for Group Agent Memory over shared contexts, as illustrated in Figure~\ref{fig:figure2}.

\subsection{Interaction over Shared Contexts}
In multi-user shared contexts, fixed memory organization and strategies struggle to adapt to evolving group information and task requirements. To address this limitation, EvoGroup enables self-evolution by treating the harness itself as the object of evolution. 

Specifically, EvoGroup starts from a minimal modular harness $H_0$. The initial harness is equipped with only six basic tools: list\_files, read\_file, grep\_search, write\_file, query\_records and create\_file. From this cold start, EvoGroup repeatedly cycles through four stages: \emph{interaction}, \emph{trajectory selection}, \emph{analysis}, and \emph{revision}, allowing the harness to progressively accumulate memories and skills from interaction experience and thereby producing a sequence of increasingly capable harnesses:
\begin{equation}
    H_0,H_1,\ldots,H_T.
\end{equation}

At iteration $t$, the Group Agent interacts with a shared group context and solves tasks using the current harness $H_t$. For a query $q_i$, the interaction produces an answer $a_i$ and a trajectory
\begin{equation}
    \tau_i = \bigl(q_i, o_i^{1}, u_i^{1}, \ldots, o_i^{L_i}, u_i^{L_i}, a_i\bigr),
\end{equation}
where $o_i^{\ell}$ denotes the information observed at step $\ell$ and $u_i^{\ell}$ denotes the corresponding reasoning, communication, or tool-use action. The trajectory therefore records not only the final response, but also the intermediate decisions through which the Group Agent uses the components of $H_t$.

Along with its answer, the Group Agent reports a subjective confidence score $c_i \in [0,1]$. We use a threshold $\gamma$ (set to $0.5$ in Figure~\ref{fig:figure2}) to distinguish confident from unconfident responses. If $c_i \leq \gamma$, EvoGroup asks the Group Agent to revisit its reasoning and explain the source of its uncertainty. This self-reflection is generated before any supervision is revealed: neither the ground-truth answer nor the official evaluation result is provided to the agent. Consequently, the reflection captures the agent's own diagnosis.

\subsection{Confidence-Guided Trajectory Selection}
\label{sec:confidence-guided-experience-selection}
During continuous interaction, different trajectories provide different evolutionary signals for self-evolution. We therefore combine the Agent's confidence with official evaluation results to select informative trajectories.

After the interaction, an official evaluator supplies an external correctness signal $y_i \in \{0,1\}$ for the answer $a_i$. Combining $y_i$ with the confidence indicator $z_i=\mathbf{1}[c_i>\gamma]$ partitions the collected trajectories into four groups:
\begin{equation}
\begin{aligned}
    \mathcal{D}_{\mathrm{CC}} &= \{\tau_i \mid z_i=1,\; y_i=1\},\\
    \mathcal{D}_{\mathrm{CW}} &= \{\tau_i \mid z_i=1,\; y_i=0\},\\
    \mathcal{D}_{\mathrm{UC}} &= \{\tau_i \mid z_i=0,\; y_i=1\},\\
    \mathcal{D}_{\mathrm{UW}} &= \{\tau_i \mid z_i=0,\; y_i=0\}.
\end{aligned}
\end{equation}
The confident-and-correct subset $\mathcal{D}_{\mathrm{CC}}$ represents routine successful behavior and is not prioritized for diagnosis. EvoGroup instead selects
\begin{equation}
    \mathcal{D}_{\mathrm{sel}}
    = \mathcal{D}_{\mathrm{CW}} \cup \mathcal{D}_{\mathrm{UC}} \cup \mathcal{D}_{\mathrm{UW}},
\end{equation}
which covers three complementary signals. Confident but incorrect trajectories reveal undetected reasoning failures or answer biases; unconfident but correct trajectories indicate fragile successes; and unconfident and incorrect trajectories capture failures that the Group Agent can partially recognize.

\subsection{Trajectory Analysis}
Agent trajectories contain numerous tool calls, observations, and shared-context details, making it difficult to locate the evidence responsible for a failure. EvoGroup addresses this challenge at two levels: it first uses \emph{Adaptive Progressive Disclosure} (APD) to inspect and attribute individual trajectories on demand, and then uses bucketed analysis to aggregate recurring issues across trajectories into high-level findings.

\noindent\textbf{APD Analysis.}
Rather than loading an entire, potentially long interaction into context at once, APD initially keeps the full trajectory hidden and exposes only a compact structural view. The Analysis Agent first locates a potentially problematic stage and then jumps to the corresponding local segment, requesting additional observations, actions, tool results, or communication records only when needed. This on-demand access focuses the analysis on diagnostically relevant evidence while reducing distraction from unrelated parts of the trajectory.

For each $\tau_i\in\mathcal{D}_{\mathrm{sel}}$, the Analysis Agent identifies the stage at which the behavior deviates from a reliable solution process, attributes the observed failure or uncertainty to its most likely cause, and produces a detailed rationale $r_i^{\mathrm{det}}$. EvoGroup then compresses the detailed diagnosis into a condensed rationale:
\begin{equation}
    r_i^{\mathrm{con}}
    =
    \operatorname{Compress}\!\left(r_i^{\mathrm{det}}\right),
\end{equation}
which retains the failure location, attributed cause, and actionable implication while removing instance-specific details that are unnecessary for cross-trajectory comparison.

\noindent\textbf{Bucketed Analysis.}
To discover recurring issues beyond individual cases, EvoGroup organizes condensed rationales according to the semantic types of their queries. These types are not predefined. Instead, the Analysis Agent infers a type label $g_i$ from each query, for example, user profiling or temporal reasoning, and assigns analyses with the same inferred type to a shared bucket:
\begin{equation}
    \mathcal{B}_{g}
    =
    \{r_i^{\mathrm{con}}\mid g_i=g\}.
\end{equation}
The Analysis Agent first analyzes evidence within each bucket to identify type-specific issues, and then compares the resulting issues across buckets to derive high-level findings. The resulting condensed rationales and high-level findings $F_t$ summarize recurring sources of error and brittle success in the current Harness, and provide actionable evidence for subsequent harness revision.

\subsection{Harness Revision}
To enable more comprehensive and effective harness revision, we organize updates into four complementary interfaces: representation, operation, intervention, and policy, each targeting a distinct aspect of harness behavior to support continuous self-evolution from interaction.

\noindent\textbf{Representation.}
This interface modifies how information is represented, organized, and linked within the Harness, without changing selection strategies. Updates include restructuring memory units, entity relations, working memory, and the context provided to the underlying model.
\noindent\textbf{Operation.}
This interface adds or modifies executable capabilities and their implementations, such as introducing query tools or revising retrieval algorithms. It determines what the harness can execute and how each operation is implemented.
\noindent\textbf{ Intervention.}
This interface introduces checks, transformations, or constraints at execution boundaries, such as validating tool arguments before calls, normalizing returned results, and verifying outputs before returning them as final answers.
\noindent\textbf{Policy.}
This interface modifies the rules governing operation selection, timing, and ordering. Updates include adjusting prompts, skill invocation rules, retrieval parameters, search strategies, and operation composition and scheduling.
Based on this distinction, the Evolve Agent selects the interface or interfaces that directly address each diagnosed issue.

\section{Experiments}
\label{sec:experiments}
\subsection{Setup}

\textbf{Benchmark \& Baseline Methods.}
We use multi-party conversation benchmarks to evaluate the memory capabilities of group-chat agents. EverMemBench \citep{hu2026evaluatinglonghorizonmemorymultiparty} is divided using stratified sampling, with 30\% used as the held-in set for harness evolution and the remaining 70\% used as the held-out set for evaluating the evolved harness.
To evaluate cross-benchmark transfer, we then freeze the evolved harness and evaluate it on GroupMemBench  \citep{yang2026groupmembench}.
We compare EvoGroup with structured-memory and self-evolving memory baselines, as well as human-designed agent Harnesses, including (1) FullContext, (2) Mem0 \citep{chhikara2025mem0}, (3) MemOS \citep{kang2025memoryosaiagent}, (4) A-MEM \citep{xu2025amem}, (5) MemRL \citep{zhang2026memrlselfevolvingagentsruntime}, (6) Codex \citep{openai2025codex}, and (7) Claude Code \citep{anthropic2025claudecode}.

\noindent\textbf{Models.}
For the evolution iterations and frozen transfer experiments (Sections~\ref{sec:main_results}, the Group Agent, Analysis Agent, and Evolve Agent use the same base model within each run. We evaluate three model configurations: GPT-5.5 \citep{openai2026GPT55} with the non-thinking reasoning setting, DeepSeek-V4-Flash \citep{Deepseek2026v4flash}, and GLM-5.1 \citep{glm5team2026glm5vibecodingagentic}.

% Background colors for answer-model headers
\definecolor{EverGPTBg}{HTML}{E5EFE9}
\definecolor{EverDeepSeekBg}{HTML}{E6EDFC}
\definecolor{EverGLMBg}{HTML}{ECECEF}

\def\evergptmodelicon{%
  \raisebox{-0.20ex}{\includegraphics[height=1.55ex]{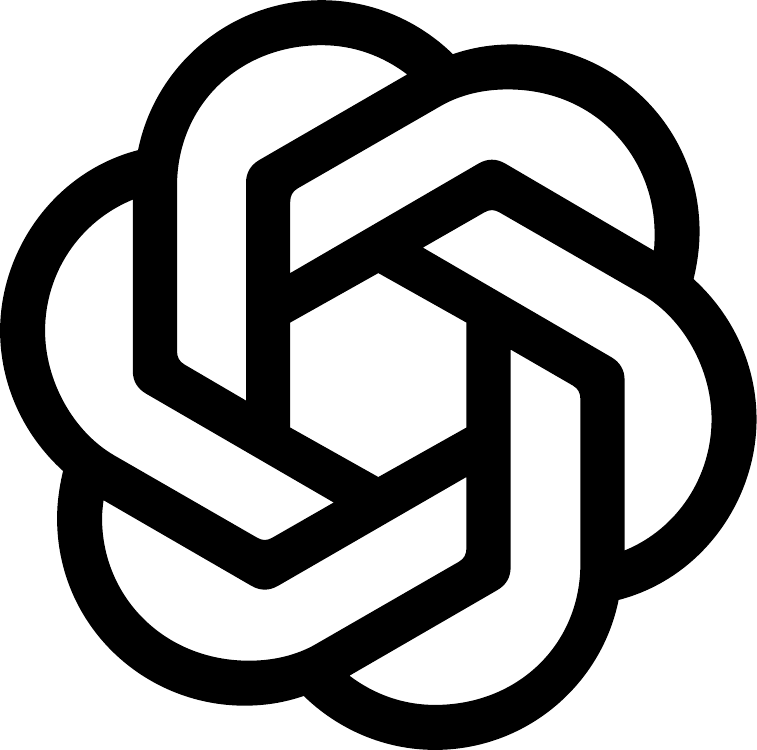}}%
}
\def\everdeepseekmodelicon{%
  \raisebox{-0.20ex}{\includegraphics[height=1.55ex]{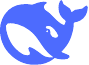}}%
}
\def\everglmmodelicon{%
  \raisebox{-0.20ex}{\includegraphics[height=1.55ex]{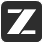}}%
}

\begin{table*}[t]
\centering
\footnotesize

% ==================== Caption spacing ====================
\setlength{\abovecaptionskip}{2pt}
\setlength{\belowcaptionskip}{2pt}

\caption{\textbf{Performance on EverMemBench.}
All category scores are aligned to the 1,680-question held-out counts (Single=149, Multi=174, Temp=210, Const=282, Proact=299, Update=188, Style=123, Skill=118, Role=137).
Baseline entries are count-aligned projections from the reported full-benchmark category accuracies, not rerun results on the held-out question IDs.
Average is recomputed from the aligned integer correct counts.
\textbf{Bold} marks the best aligned result within each answer-model block, excluding FullContext.}
\label{tab:evermembench-main}

% ==================== Compact vertical spacing ====================
\setlength{\aboverulesep}{0.15ex}
\setlength{\belowrulesep}{0.15ex}
\setlength{\cmidrulesep}{0.10ex}
\setlength{\defaultaddspace}{0pt}

\renewcommand{\arraystretch}{0.82}
\setlength{\tabcolsep}{2.7pt}

% Model header height
\def\evermodelbarstrut{\rule[-0.45ex]{0pt}{1.75ex}}

% Group label height and spacing
\def\evergrouplabelstrut{\rule{0pt}{1.35ex}}
\def\evergrouplabelsep{0pt}

\hfuzz=3pt

\ifdefined\dashlinedash
  \setlength\dashlinedash{0.6pt}
  \setlength\dashlinegap{1pt}
\fi

\begin{tabularx}{\dimexpr\textwidth-3pt\relax}{
  @{}l*{10}{>{\centering\arraybackslash}X}@{}
}
\toprule

\multirow[c]{2}{*}{%
  \raisebox{-0.25ex}{\hspace{2.5pt}Method}%
}
&
\multicolumn{3}{c}{%
  \makebox[0pt][c]{Fine-Grained Recall}%
}
&
\multicolumn{3}{c}{%
  \makebox[0pt][c]{Memory Awareness}%
}
&
\multicolumn{3}{c}{%
  \makebox[0pt][c]{Profile Understanding}%
}
&
\multirow[c]{2}{*}{%
  \makebox[\linewidth][c]{%
    \raisebox{-0.25ex}{Average}%
  }%
}
\\

\cmidrule(l{1pt}r{1pt}){2-4}
\cmidrule(l{1pt}r{1pt}){5-7}
\cmidrule(l{1pt}r{1pt}){8-10}

& \makebox[\linewidth][c]{\textit{Single}}
& \makebox[\linewidth][c]{\textit{Multi}}
& \makebox[\linewidth][c]{\textit{Temp}}
& \makebox[\linewidth][c]{\textit{Const}}
& \makebox[\linewidth][c]{\textit{Proact}}
& \makebox[\linewidth][c]{\textit{Update}}
& \makebox[\linewidth][c]{\textit{Style}}
& \makebox[\linewidth][c]{\textit{Skill}}
& \makebox[\linewidth][c]{\textit{Role}}
& \\
\midrule

% ==================== GPT-5.5 ====================
\multicolumn{11}{>{\columncolor{EverGPTBg}}c}{%
  \evermodelbarstrut
  \evergptmodelicon\hspace{0.35em}\textbf{\textit{GPT-5.5}}%
} \\
\midrule

\hspace{2.5pt}FullContext
& 97.32 & 4.02 & 18.10 & 92.20 & 84.62
& 100.00 & 47.97 & 63.56 & 65.69 & 66.37 \\

\evergroupspacedash
\multicolumn{11}{l}{%
  \evergrouplabelstrut
  \hspace{4.5pt}\textit{Structured Memory}
} \\[-0.4ex]

\hspace{7.5pt}Mem0
& 88.59 & 4.02 & 10.00 & 85.11 & 77.59
& 75.53 & 17.07 & 63.56 & 57.66 & 56.49 \\

\hspace{7.5pt}MemOS
& 71.14 & 1.72 & 5.24 & 68.79 & 62.21
& 65.96 & 19.51 & \textbf{69.49} & 70.80 & 49.23 \\

\evergroupspacedash
\multicolumn{11}{l}{%
  \evergrouplabelstrut
  \hspace{4.5pt}\textit{Self-Evolving Memory}
} \\[-0.4ex]

\hspace{7.5pt}A-MEM
& 90.60 & 4.02 & 21.90 & 89.01 & 88.29
& 75.53 & 27.64 & 57.63 & 65.69 & 61.73 \\

\hspace{7.5pt}MemRL
& 90.60 & 5.75 & 21.90 & 90.78 & 82.27
& 77.66 & 30.08 & 54.24 & 68.61 & 61.55 \\

\evergroupspacedash
\multicolumn{11}{l}{%
  \evergrouplabelstrut
  \hspace{4.5pt}\textit{Human-Designed Harness}
} \\[-0.4ex]

\hspace{7.5pt}Codex
& 93.96 & \textbf{76.44} & 75.24 & 96.10 & \textbf{91.30}
& 97.34 & \textbf{78.86} & 67.80 & \textbf{75.18}
& \textbf{85.60} \\

\hspace{7.5pt}Claude Code
& 93.29 & 74.71 & 73.81 & 95.74 & 90.30
& 96.81 & 78.05 & 66.10 & 74.45 & 84.64 \\

\evergroupspacedash
\multicolumn{11}{l}{%
  \evergrouplabelstrut
  \hspace{4.5pt}\textit{EvoGroup}
} \\[-0.4ex]

\hspace{7.5pt}$H_0$
& 95.30 & 8.05 & 51.90 & 90.43 & 88.29
& 95.74 & 36.59 & 45.76 & 62.04 & 68.33 \\

\hspace{7.5pt}$H_{*}$
& \textbf{97.32} & 56.90 & \textbf{85.24} & \textbf{96.81}
& 90.97 & \textbf{97.87} & 74.80 & 64.41 & 72.99 & 84.52 \\

\midrule

% ==================== DeepSeek-V4-Flash ====================
\multicolumn{11}{>{\columncolor{EverDeepSeekBg}}c}{%
  \evermodelbarstrut
  \everdeepseekmodelicon\hspace{0.35em}
  \textbf{\textit{DeepSeek-V4-Flash}}%
} \\
\midrule

\hspace{2.5pt}FullContext
& 95.30 & 5.75 & 11.90 & 50.35 & 49.50
& 100.00 & 19.51 & 18.64 & 23.36 & 43.63 \\

\evergroupspacedash
\multicolumn{11}{l}{%
  \evergrouplabelstrut
  \hspace{4.5pt}\textit{Structured Memory}
} \\[-0.4ex]

\hspace{7.5pt}Mem0
& 81.21 & 4.02 & 6.67 & 55.67 & 57.53
& 75.53 & 21.95 & 27.12 & 45.26 & 43.69 \\

\hspace{7.5pt}MemOS
& 64.43 & 4.02 & 13.33 & 47.16 & 58.86
& 65.96 & 15.45 & 18.64 & 39.42 & 39.23 \\

\evergroupspacedash
\multicolumn{11}{l}{%
  \evergrouplabelstrut
  \hspace{4.5pt}\textit{Self-Evolving Memory}
} \\[-0.4ex]

\hspace{7.5pt}A-MEM
& 88.59 & 4.02 & 18.10 & 63.48 & 56.52
& 77.66 & 33.33 & 27.12 & 34.31 & 47.08 \\

\hspace{7.5pt}MemRL
& 93.29 & 0.00 & 0.00 & 81.21 & 83.61
& 77.66 & 23.58 & 48.31 & 65.69 & 55.95 \\

\evergroupspacedash
\multicolumn{11}{l}{%
  \evergrouplabelstrut
  \hspace{4.5pt}\textit{Human-Designed Harness}
} \\[-0.4ex]

\hspace{7.5pt}Codex
& 95.30 & 82.18 & 83.33 & 93.62 & 90.64
& 92.02 & 78.05 & 50.00 & 47.45 & 82.62 \\

\hspace{7.5pt}Claude Code
& 97.32 & \textbf{83.91} & 86.67 & \textbf{98.58}
& \textbf{92.98} & 93.62 & 84.55 & 42.37 & 60.58 & 85.83 \\

\evergroupspacedash
\multicolumn{11}{l}{%
  \evergrouplabelstrut
  \hspace{4.5pt}\textit{EvoGroup}
} \\[-0.4ex]

\hspace{7.5pt}$H_0$
& 97.99 & 28.16 & 31.90 & 89.36 & \textbf{92.98}
& 92.55 & 67.48 & 51.69 & 61.31 & 71.07 \\

\hspace{7.5pt}$H_{*}$
& \textbf{100.00} & 78.16 & \textbf{90.00} & 97.52
& 90.64 & \textbf{98.40} & \textbf{97.56} & \textbf{66.10}
& \textbf{74.45} & \textbf{89.58} \\

\midrule

% ==================== GLM-5.1 ====================
\multicolumn{11}{>{\columncolor{EverGLMBg}}c}{%
  \evermodelbarstrut
  \everglmmodelicon\hspace{0.35em}\textbf{\textit{GLM-5.1}}%
} \\
\midrule

\hspace{2.5pt}FullContext
& -- & -- & -- & -- & -- & -- & -- & -- & -- & -- \\

\evergroupspacedash
\multicolumn{11}{l}{%
  \evergrouplabelstrut
  \hspace{4.5pt}\textit{Structured Memory}
} \\[-0.4ex]

\hspace{7.5pt}Mem0
& 88.59 & 4.02 & 10.00 & 67.38 & 68.23
& 79.26 & 19.51 & 48.31 & 70.80 & 52.44 \\

\hspace{7.5pt}MemOS
& 73.83 & 5.75 & 3.33 & 58.16 & 57.53
& 86.70 & 13.82 & 48.31 & 65.69 & 47.02 \\

\evergroupspacedash
\multicolumn{11}{l}{%
  \evergrouplabelstrut
  \hspace{4.5pt}\textit{Self-Evolving Memory}
} \\[-0.4ex]

\hspace{7.5pt}A-MEM
& 88.59 & 4.02 & 18.10 & 76.95 & 83.61
& 79.26 & 34.96 & 42.37 & \textbf{73.72} & 58.75 \\

\hspace{7.5pt}MemRL
& 93.29 & 4.02 & 20.00 & 75.89 & 78.93
& 81.38 & 33.33 & 36.44 & 65.69 & 57.44 \\

\evergroupspacedash
\multicolumn{11}{l}{%
  \evergrouplabelstrut
  \hspace{4.5pt}\textit{Human-Designed Harness}
} \\[-0.4ex]

\hspace{7.5pt}Codex
& 95.30 & \textbf{67.82} & 63.33 & \textbf{96.10}
& 81.27 & 88.83 & 80.49 & 41.53 & 63.50 & 77.92 \\

\hspace{7.5pt}Claude Code
& 95.97 & 55.75 & 65.24 & 93.26 & 86.96
& \textbf{100.00} & 85.37 & 20.34 & 50.36 & 76.55 \\

\evergroupspacedash
\multicolumn{11}{l}{%
  \evergrouplabelstrut
  \hspace{4.5pt}\textit{EvoGroup}
} \\[-0.4ex]

\hspace{7.5pt}$H_0$
& 94.63 & 24.71 & 28.10 & 86.88 & 89.97
& 89.36 & 60.98 & 56.78 & 58.39 & 68.27 \\

\hspace{7.5pt}$H_{*}$
& \textbf{97.99} & 66.09 & \textbf{82.86} & 90.07
& \textbf{90.30} & 81.38 & \textbf{90.24} & \textbf{72.03}
& 70.80 & \textbf{83.63} \\

\bottomrule
\end{tabularx}
\end{table*}

\noindent\textbf{Metric.}
We evaluate task effectiveness using \textbf{pass@1}. Each task is attempted once per evaluation run and is considered successful if it passes the official verifier. The final score is the proportion of successfully completed tasks. Trials terminated by system-level failures, such as timeouts, crashes, or missing outputs, are counted as unsuccessful. Detailed experimental settings are provided in Appendix~\ref{app:experimental-setup}.

\subsection{Main Results}
\label{sec:main_results}
As shown in Table~\ref{tab:evermembench-main}, we evaluate EvoGroup on GPT-5.5, DeepSeek-V4-Flash, and GLM-5.1. After six self-evolution iterations, EvoGroup identifies the best-performing harness configuration, denoted by \(H_{*}\). On the held-out dataset, compared with the initial configuration \(H_0\), \(H_{*}\) improves the average pass rate by 16.2, 18.5, and 15.4 pp on the three models, respectively, demonstrating EvoGroup's ability to consistently discover substantially better configurations. Meanwhile, EvoGroup outperforms the Structured Memory and Self-Evolving Memory baselines, as well as Claude Code and Codex, across all three backbones, except that its performance on GPT-5.5 is slightly below both Codex and Claude Code. However, since the evolution process is not limited to the number of iterations evaluated in our experiments, further iterations may yield additional performance gains.
These results demonstrate that \textbf{EvoGroup consistently discovers substantially stronger memory harness configurations across LLM backbones with different capabilities.}

\noindent\textbf{Generalization beyond the evolution benchmark.}
To evaluate whether the evolved harness generalizes beyond the distribution used for evolution, we freeze the harness evolved on EverMemBench and directly evaluate it on the unseen GroupMemBench without further adaptation.
As shown in Table~\ref{tab:groupmembench-main}, on GPT-5.5, the evolved harness achieves the highest average pass rate of 63.89\%, outperforming the initial harness by 2.01 points and the strongest human-designed harness by 8.46 points.
On DeepSeek-V4-Flash, it achieves 71.14\%, remaining within 2.95 points of the initial harness and 5.37 points of Codex, while outperforming the strongest memory baseline by 28.46 points.
These results show that \textbf{EvoG can transfer to an unseen benchmark while maintaining strong performance, although some performance degradation is observed with DeepSeek-V4-Flash.}

\begin{figure}[t]
    \centering
    \includegraphics[width=\columnwidth]{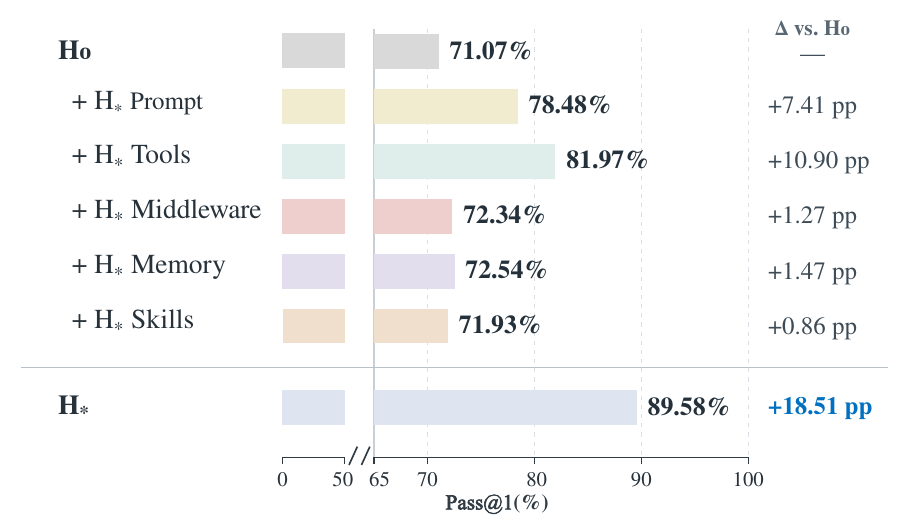}
    \caption{
    \textbf{Component ablation} on the held-out
    EverMemBench split using DeepSeek-V4-Flash. Each variant uses one
    evolved component, while \(H_{*}\) uses all five.
    \(\Delta\) indicates the gains over \(H_0\).
    }
    \label{fig:componen_ablation}
    \vspace{-0.5em}
\end{figure}

% Process ablation
\begin{figure}[t]
    \centering
    \includegraphics[width=\columnwidth]{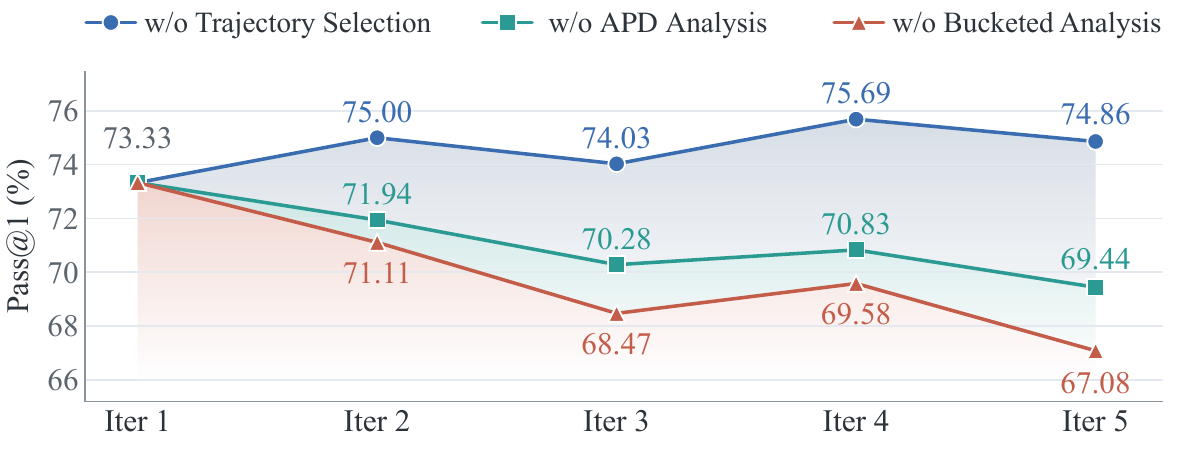}
    \caption{
    \textbf{Process ablation} on the held-in
    EverMemBench split using DeepSeek-V4-Flash.
    Pass@1 under three ablation
    settings: w/o Trajectory Selection, w/o APD Analysis, and
    w/o Bucketed Analysis.
    }
    \label{fig:self-evolution-zhexian-ablation}
    \vspace{-0.5em}
\end{figure}

\subsection{Ablation Study}
\noindent\textbf{Component-level ablations.}
We evaluate the contribution of each harness component by replacing one component of the initial harness with its counterpart from the fully evolved harness, while keeping all other components unchanged.
We consider Prompt, Tools, Middleware, Memory, and Skills separately.
As shown in Figure~\ref{fig:componen_ablation}, Tools and Prompt provide the largest gains, improving accuracy by 10.90 and 7.41 pp, respectively.
Middleware, Memory, and Skills yield smaller gains of 1.27, 1.47, and 0.86 points.
Using all evolved components achieves 89.58\% accuracy, an 18.51 pp improvement over the initial harness at 71.07\%.
These results show that EvoGroup improves the harness through multiple components rather than prompt optimization alone.

\noindent\textbf{Process-level ablations.}
We further examine three mechanisms in the self-evolution loop using DeepSeek-V4-Flash.
All variants use the same cold-start harness configuration, evolution set, and token budget.
As shown in Figure~\ref{fig:self-evolution-zhexian-ablation}, removing trajectory selection increases Pass@1 from 73.33\% to 74.86\%, with a peak of 75.69\%.
Thus, analyzing all trajectories can support self-evolution, but requires more analysis time and greater token consumption.

Replacing APD with full-trace analysis yields a final Pass@1 of 69.44\%, below its initial score. This suggests that, in shared-context task settings, directly analyzing complete trajectories may introduce irrelevant information and noise, whereas APD enables the Analysis Agent to focus on diagnostically relevant evidence and produce more effective evolution signals.

Removing bucketed analysis produces the lowest final Pass@1 of 67.08\%.
Without cross-trajectory aggregation, individual diagnoses may miss recurring failure patterns, resulting in more instance-specific and less reliable harness updates.

\begin{table*}[t]
\centering
\caption{
\textbf{Capability-transition analysis of self-evolution on the held-in EverMemBench}
under (a) DeepSeek-V4-Flash and (b) GPT-5.5.
Fail$\rightarrow$Pass and Pass$\rightarrow$Fail denote questions from
incorrect to correct and from correct to incorrect, respectively.
Always$\rightarrow$Pass and Always$\rightarrow$Fail denote questions that
remain consistently correct or incorrect across consecutive harnesses.
Retention is the fraction of previously correct questions that remain correct,
while Regression is the fraction that become incorrect in the next harness.
}
\label{tab:evogroup-transition-comparison}

\vspace{-6pt}

\begin{minipage}[t]{0.495\textwidth}
\centering
{\scriptsize (a) DeepSeek-V4-Flash}
\vspace{0pt}
\scriptsize
\setlength{\tabcolsep}{1.5pt}
\renewcommand{\arraystretch}{1.35}

\resizebox{\linewidth}{!}{%
\begin{tabular}{
    >{\centering\arraybackslash}p{1.2cm}
    >{\centering\arraybackslash}p{1.2cm}
    >{\centering\arraybackslash}p{1.2cm}
    >{\centering\arraybackslash}p{1.5cm}
    >{\centering\arraybackslash}p{1.5cm}
    >{\centering\arraybackslash}p{1.3cm}
    >{\centering\arraybackslash}p{1.3cm}
}
\toprule
Transition &
Fail$\rightarrow$Pass &
Pass$\rightarrow$Fail &
Always$\rightarrow$Pass &
Always$\rightarrow$Fail &
Retention $\uparrow$ &
Regression $\downarrow$ \\
\midrule

$H_0 \rightarrow H_1$ &
\EvoCellBar{EvoBlue}{1.000}{124} &
\EvoCellBar{EvoBlue}{0.642}{34} &
\EvoCellBar{EvoBlue}{0.807}{469} &
\EvoCellBar{EvoBlue}{0.959}{93} &
\EvoCellBar{EvoBlue}{0.9324}{93.24\%} &
\EvoCellBar{EvoBlue}{0.0676}{6.76\%} \\

$H_1 \rightarrow H_2$ &
\EvoCellBar{EvoGreen}{0.589}{73} &
\EvoCellBar{EvoGreen}{0.604}{32} &
\EvoCellBar{EvoGreen}{0.966}{561} &
\EvoCellBar{EvoGreen}{0.557}{54} &
\EvoCellBar{EvoGreen}{0.9460}{94.60\%} &
\EvoCellBar{EvoGreen}{0.0540}{5.40\%} \\

$H_2 \rightarrow H_3$ &
\EvoCellBar{EvoRed}{0.306}{38} &
\EvoCellBar{EvoRed}{1.000}{53} &
\EvoCellBar{EvoRed}{1.000}{581} &
\EvoCellBar{EvoRed}{0.495}{48} &
\EvoCellBar{EvoRed}{0.9164}{91.64\%} &
\EvoCellBar{EvoRed}{0.0836}{8.36\%} \\

$H_3 \rightarrow H_4$ &
\EvoCellBar{EvoOrange}{0.363}{45} &
\EvoCellBar{EvoOrange}{0.925}{49} &
\EvoCellBar{EvoOrange}{0.981}{570} &
\EvoCellBar{EvoOrange}{0.577}{56} &
\EvoCellBar{EvoOrange}{0.9208}{92.08\%} &
\EvoCellBar{EvoOrange}{0.0792}{7.92\%} \\

$H_4 \rightarrow H_5$ &
\EvoCellBar{EvoPurple}{0.427}{53} &
\EvoCellBar{EvoPurple}{0.830}{44} &
\EvoCellBar{EvoPurple}{0.983}{571} &
\EvoCellBar{EvoPurple}{0.536}{52} &
\EvoCellBar{EvoPurple}{0.9285}{92.85\%} &
\EvoCellBar{EvoPurple}{0.0715}{7.15\%} \\

\bottomrule
\end{tabular}%
}
\end{minipage}
\hfill
\begin{minipage}[t]{0.495\textwidth}
\centering
{\scriptsize (b) GPT-5.5}
\vspace{0pt}
\scriptsize
\setlength{\tabcolsep}{1.5pt}
\renewcommand{\arraystretch}{1.35}

\resizebox{\linewidth}{!}{%
\begin{tabular}{
    >{\centering\arraybackslash}p{1.2cm}
    >{\centering\arraybackslash}p{1.2cm}
    >{\centering\arraybackslash}p{1.2cm}
    >{\centering\arraybackslash}p{1.5cm}
    >{\centering\arraybackslash}p{1.5cm}
    >{\centering\arraybackslash}p{1.3cm}
    >{\centering\arraybackslash}p{1.3cm}
}
\toprule
Transition &
Fail$\rightarrow$Pass &
Pass$\rightarrow$Fail &
Always$\rightarrow$Pass &
Always$\rightarrow$Fail &
Retention $\uparrow$ &
Regression $\downarrow$ \\
\midrule

$H_0 \rightarrow H_1$ &
\EvoCellBar{EvoBlue}{0.653}{81} &
\EvoCellBar{EvoBlue}{0.679}{36} &
\EvoCellBar{EvoBlue}{0.871}{506} &
\EvoCellBar{EvoBlue}{1.000}{97} &
\EvoCellBar{EvoBlue}{0.9340}{93.4\%} &
\EvoCellBar{EvoBlue}{0.0660}{6.6\%} \\

$H_1 \rightarrow H_2$ &
\EvoCellBar{EvoGreen}{0.315}{39} &
\EvoCellBar{EvoGreen}{0.792}{42} &
\EvoCellBar{EvoGreen}{0.938}{545} &
\EvoCellBar{EvoGreen}{0.969}{94} &
\EvoCellBar{EvoGreen}{0.9280}{92.8\%} &
\EvoCellBar{EvoGreen}{0.0720}{7.2\%} \\

$H_2 \rightarrow H_3$ &
\EvoCellBar{EvoRed}{0.339}{42} &
\EvoCellBar{EvoRed}{0.830}{44} &
\EvoCellBar{EvoRed}{0.929}{540} &
\EvoCellBar{EvoRed}{0.969}{94} &
\EvoCellBar{EvoRed}{0.9250}{92.5\%} &
\EvoCellBar{EvoRed}{0.0750}{7.5\%} \\

$H_3 \rightarrow H_4$ &
\EvoCellBar{EvoOrange}{0.411}{51} &
\EvoCellBar{EvoOrange}{0.623}{33} &
\EvoCellBar{EvoOrange}{0.945}{549} &
\EvoCellBar{EvoOrange}{0.897}{87} &
\EvoCellBar{EvoOrange}{0.9430}{94.3\%} &
\EvoCellBar{EvoOrange}{0.0570}{5.7\%} \\

$H_4 \rightarrow H_5$ &
\EvoCellBar{EvoPurple}{0.379}{47} &
\EvoCellBar{EvoPurple}{0.792}{42} &
\EvoCellBar{EvoPurple}{0.960}{558} &
\EvoCellBar{EvoPurple}{0.753}{73} &
\EvoCellBar{EvoPurple}{0.9300}{93.0\%} &
\EvoCellBar{EvoPurple}{0.0700}{7.0\%} \\

\bottomrule
\end{tabular}%
}
\end{minipage}
\end{table*}

\begin{table}[htpb]
\centering
\small
\setlength{\tabcolsep}{3pt}
\renewcommand{\arraystretch}{1.14}
\caption{\textbf{Effectiveness of Adaptive Progressive Disclosure (APD) for trajectory analysis.}
We compare APD Analysis with Full Trace Analysis across self-evolution iterations
in terms of analysis time and token consumption.
APD consistently reduces both analysis time and token consumption compared with
Full Trace Analysis, demonstrating its
effectiveness in improving trajectory-analysis efficiency.
Time is reported in minutes (min), and token usage in millions of tokens (M).}
\label{tab:apd-efficiency}

\resizebox{\columnwidth}{!}{%
\begin{tabular}{lcccc}
\toprule
\multirow{2}{*}{\textbf{Iteration}} &
\multicolumn{2}{c}{\textbf{Time (min)}} &
\multicolumn{2}{c}{\textbf{Tokens (M)}} \\
\cmidrule(lr){2-3}\cmidrule(lr){4-5}
& \textit{APD} & \textit{Full} & \textit{APD} & \textit{Full} \\
\midrule
\iterkey{tabPurple}{Iteration 1} &
\metricbar{1.000}{tabPurple}{64.6} &
\metricbar{1.000}{tabPurple}{110.8} &
\metricbar{1.000}{tabPurple}{54.58} &
\metricbar{1.000}{tabPurple}{82.09} \\

\iterkey{tabBlue}{Iteration 2} &
\metricbar{0.404}{tabBlue}{26.1} &
\metricbar{0.388}{tabBlue}{43.0} &
\metricbar{0.611}{tabBlue}{33.34} &
\metricbar{0.811}{tabBlue}{66.55} \\

\iterkey{tabGreen}{Iteration 3} &
\metricbar{0.333}{tabGreen}{21.5} &
\metricbar{0.342}{tabGreen}{37.9} &
\metricbar{0.405}{tabGreen}{22.08} &
\metricbar{0.416}{tabGreen}{34.18} \\

\iterkey{tabOrange}{Iteration 4} &
\metricbar{0.274}{tabOrange}{17.7} &
\metricbar{0.272}{tabOrange}{30.1} &
\metricbar{0.521}{tabOrange}{28.46} &
\metricbar{0.683}{tabOrange}{56.12} \\

\iterkey{tabMagenta}{Iteration 5} &
\metricbar{0.385}{tabMagenta}{24.9} &
\metricbar{0.429}{tabMagenta}{47.5} &
\metricbar{0.438}{tabMagenta}{23.88} &
\metricbar{0.471}{tabMagenta}{38.64} \\
\bottomrule
\end{tabular}%
}
\vspace{-1.0em}
\end{table}

\subsection{Analysis of Capability Transition}
\label{sec:capability-transition}
To examine how model capabilities evolve during self-evolution, we analyze task-level transitions between consecutive evolution rounds. As shown in Table~\ref{tab:evogroup-transition-comparison}, EvoGroup generally acquires new capabilities while retaining existing ones, although capability stability varies across models and evolution rounds. DeepSeek-V4-Flash maintains a retention rate above 91\%, with regression rates ranging from 5.40\% to 8.36\%. GPT-5.5 also exhibits stable capability retention, with retention rates ranging from 92.5\% to 94.3\% and regression rates ranging from 5.7\% to 7.5\%.
These results demonstrate that EvoGroup can improve task-solving capabilities across different LLM backbones while preserving most existing capabilities and maintaining relatively low regression rates throughout the evolution process.

\subsection{Analysis of Confidence and Bias}
We examine whether self-evolution improves both task performance and agent confidence by grouping trajectories into the four types defined in Section~\ref{sec:confidence-guided-experience-selection}. As shown in Figure~\ref{fig:Confidence}, for DeepSeek-V4-Flash, the proportion of confident-and-correct (CC) trajectories increases by 15.83\%, while that of confident-but-wrong (CW) trajectories decreases by 6.25 \%, indicating improvements in both task performance and confidence calibration, despite some fluctuations across intermediate iterations. For GPT-5.5, the proportion of CC trajectories increases by 10.62\%, while that of CW trajectories decreases by 6.04\%. Likewise, this shift indicates improved task performance and a more favorable confidence–correctness distribution. Overall, both models exhibit a higher proportion of confident-and-correct trajectories and a lower proportion of confident-but-incorrect trajectories after self-evolution.

\subsection{Analysis of Self-reflection}
To assess the agreement between self-reflection and trajectory-based diagnosis, we adopt three measures: localization-agreement, cause attribution-agreement, and diagnostic-evidence consistency.
Localization-agreement reaches 84.0\% for DeepSeek-V4-Flash and 74.4\% for GPT-5.5, suggesting that self-reflection can often identify where a problem likely occurs without examining the full trajectory. In contrast, cause attribution-agreement is 49.2\% and 62.0\% for the two models, while diagnostic-evidence consistency is only 7.1\% and 16.8\%, respectively.
These results indicate that self-reflection can provide a valuable yet not fully reliable localization signal before the full trajectory is read, directing the Analysis Agent toward likely problem regions and offering a more efficient, token-saving diagnostic pathway for harness self-evolution. Nevertheless, accurately verifying the root cause with sufficient evidence still relies on full trajectory analysis. Further improving the agent's autonomous error-localization ability could reduce this reliance and further accelerate the harness self-evolution loop.

\vspace{-1em}
\begin{figure}[htpb]
\centering
\includegraphics[width=\linewidth]{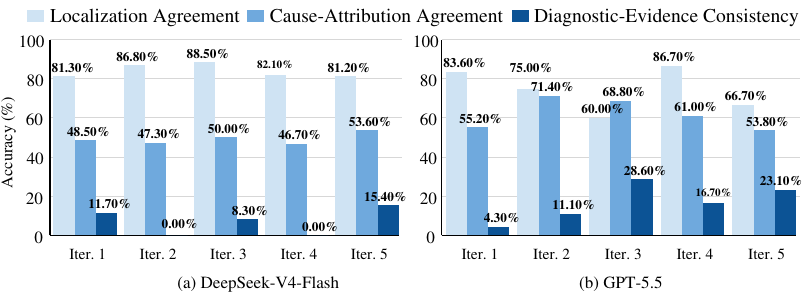}
\vspace{-1em}
\caption{\textbf{Agreement between self-reflection and trajectory-based diagnosis}. \textbf{Localization Agreement} indicates whether self-reflection correctly locates the problematic region, \textbf{Cause-Attribution Agreement} whether it identifies the underlying cause, and \textbf{Diagnostic-Evidence Consistency} whether its diagnosis is supported by trajectory evidence.}
\label{fig:self-evolution}
\vspace{-1.0em}
\end{figure}

\subsection{Analysis of Adaptive Progressive Disclosure}
We compare Adaptive Progressive Disclosure (APD) with Full Trace Analysis while keeping the analysis model, budget, concurrency, and input content fixed.
As shown in Table~\ref{tab:apd-efficiency}, APD reduces analysis time by 42.5\% and token consumption by 41.5\% on average across the 5 iterations.
Most of the reduction comes from fewer input tokens, while output length remains similar.
This shows that APD improves analysis efficiency by selectively accessing relevant trajectory segments instead of processing the full trace.

\begin{figure}[htpb]
\centering
\includegraphics[width=\linewidth]{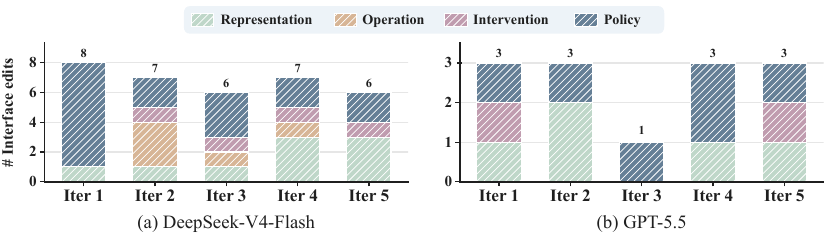}
\vspace{-1em}
\caption{\textbf{Revision-interface edits} across evolution rounds for DeepSeek-V4-Flash and GPT-5.5. The figure illustrates which revision interfaces EvoGroup relies on most during self-evolution.}
\vspace{-1.0em}
\label{fig:interface_edits}
\end{figure}

\subsection{Effectiveness of Bucketed Analysis}
Table~\ref{tab:adoption-repair} evaluates whether bucketed analysis produces useful evolution signals. Across self-evolution iterations, DeepSeek-V4-Flash achieves finding correctness rates of 71.40\%-100.00\%, adoption rates of 57.10\%-66.70\%, and repair rates of 41.40\%-59.40\%. GPT-5.5 maintains a finding correctness rate of 100.00\%, with adoption rates of 50.00\%-87.50\% and repair rates of 33.30\%-68.60\%. These results show that bucketed analysis can distill recurring task-level failures into high-level findings that guide the Evolve Agent's actions, although the resulting updates may still introduce regressions.

\subsection{Analysis of Interface Importance}
As shown in Figure~\ref{fig:interface_edits}, we analyze the use of the four revision interfaces by the Evolve Agent across evolution iterations. Policy accounts for the largest share of revisions, at 50.00\% for DeepSeek-V4-Flash and 46.15\% for GPT-5.5, followed by Representation. This may be because many problems encountered by group agents operating in shared contexts involve information retrieval and organization, as well as contextual dependencies across interactions. Both models use Intervention relatively infrequently, possibly because many problems can be addressed by adjusting policies or information representations without introducing additional runtime checks or constraints. Overall, EvoGroup primarily evolves through Policy and Representation updates, while modifications to executable capabilities and execution-boundary controls are more targeted. GPT-5.5 adopts a more conservative approach to self-evolution, whereas DeepSeek-V4-Flash makes more aggressive changes.

\section{Related Work}

\textbf{Agent Memory.}
Agent memory aims to store and retrieve interaction information to support subsequent tasks~\citep{park2023generativeagents,packer2023memgpt,zhong2024memorybank}. Existing methods typically organize memory using predefined structures (e.g., trees or graphs), combined with mechanisms such as profiling, reflection, and semantic compression~\citep{park2023generativeagents,shinn2023reflexion,packer2023memgpt,chhikara2025mem0,xu2025amem,fang2026lightmem,liu2026simplemem}. However, their memory storage and retrieval mechanisms are relatively fixed, making it difficult to adapt memory management to task requirements. Recent work has begun exploring self-evolving memory by optimizing memory operations through reinforcement learning or improving memory construction and updating~\citep{yan2025memoryr1,xu2025amem,liu2026simplemem}. However, these methods primarily target single-user.
\citet{hu2026evaluatinglonghorizonmemorymultiparty,yang2026groupmembench} have begun to investigate group-chat memory and its evaluation. Unlike single-user memory, group-chat memory must integrate information distributed across speakers, sessions, and temporal states while preserving user attribution, entity relationships, and contextual dependencies. To overcome the limitations of single-user designs and fixed memory paradigms, we propose EvoG, which enables memory mechanisms to evolve continuously based on interaction feedback.

\noindent\textbf{Agent harness.} An agent harness is a framework that organizes memory, planning, tools, and action policies around a foundation model to support agent-environment interaction.~\citep{tang2026harnesssurvey}.
Existing studies adapt harnesses through self-improvement, programmatic optimization, trajectory analysis, and code generation to improve task performance and adaptability~\citep{lin2026ahe,huang2026memoharness,chen2026harnessx,zhou2026hsi,zhang2026selfharness,ning2026codeagentharness,xu2026polar,banu2026harnessarchitecture}. However, most target coding, lacking efficient diagnostics for long interaction trajectories.
In multi-user shared contexts, evidence scattered across speakers, sessions, and temporal states creates lengthy, interconnected trajectories, making full-trace analysis costly and prone to irrelevant information. To address this challenge, we propose EvoGroup.

\section{Conclusion}
In this work, we introduce EvoGroup, a self-evolving harness for group agent memory. 
By treating the harness as an evolvable system and learning evolution signals in the shared context, EvoGroup enables Group Agents to autonomously adapt to diverse groups. 
Result demonstrates that EvoGroup improves efficiency while maintaining strong performance. 
In the future, we will explore more realistic collaborative scenarios with diverse and long-term multi-user.

% \section*{Limitations}

% This document does not cover the content requirements for ACL or any
% other specific venue.  Check the author instructions for
% information on
% maximum page lengths, the required ``Limitations'' section,
% and so on.

% \section*{Acknowledgments}

% This document has been adapted
% by Steven Bethard, Ryan Cotterell and Rui Yan
% from the instructions for earlier ACL and NAACL proceedings, including those for
% ACL 2019 by Douwe Kiela and Ivan Vuli\'{c},
% NAACL 2019 by Stephanie Lukin and Alla Roskovskaya,
% ACL 2018 by Shay Cohen, Kevin Gimpel, and Wei Lu,
% NAACL 2018 by Margaret Mitchell and Stephanie Lukin,
% Bib\TeX{} suggestions for (NA)ACL 2017/2018 from Jason Eisner,
% ACL 2017 by Dan Gildea and Min-Yen Kan,
% NAACL 2017 by Margaret Mitchell,
% ACL 2012 by Maggie Li and Michael White,
% ACL 2010 by Jing-Shin Chang and Philipp Koehn,
% ACL 2008 by Johanna D. Moore, Simone Teufel, James Allan, and Sadaoki Furui,
% ACL 2005 by Hwee Tou Ng and Kemal Oflazer,
% ACL 2002 by Eugene Charniak and Dekang Lin,
% and earlier ACL and EACL formats written by several people, including
% John Chen, Henry S. Thompson and Donald Walker.
% Additional elements were taken from the formatting instructions of the \emph{International Joint Conference on Artificial Intelligence} and the \emph{Conference on Computer Vision and Pattern Recognition}.

% % Custom bibliography entries only
\bibliography{acl_latex}

\clearpage
\appendix
\section{Additional Experimental Results}
\label{app:additional-analysis}

\def\evergptmodelicon{%
  \raisebox{-0.25ex}{\includegraphics[height=1.90ex]{figures/logos/openai.pdf}}%
}
\def\everdeepseekmodelicon{%
  \raisebox{-0.25ex}{\includegraphics[height=1.90ex]{figures/logos/deepseek.pdf}}%
}
\def\everglmmodelicon{%
  \raisebox{-0.25ex}{\includegraphics[height=1.90ex]{figures/logos/zai.pdf}}%
}

\begin{table*}[htpb]
\centering
\small

% ==================== Caption spacing ====================
\setlength{\abovecaptionskip}{2pt}
\setlength{\belowcaptionskip}{2pt}

\caption{\textbf{Performance on GroupMemBench.}
All metrics are computed using the official scripts and reported as
Pass@1 (\%). Average is the reported aggregate score.
\textbf{Bold} marks the best result within each answer-model block,
excluding FullContext. ``-'' indicates that the input exceeds the model's context limit.}
\label{tab:groupmembench-main}

% ==================== Compact vertical spacing ====================
\setlength{\aboverulesep}{0.2ex}
\setlength{\belowrulesep}{0.2ex}
\setlength{\cmidrulesep}{0.1ex}
\setlength{\defaultaddspace}{0pt}

\renewcommand{\arraystretch}{0.90}
\setlength{\tabcolsep}{3.5pt}

% Model headers
\def\evermodelbarstrut{\rule[-0.6ex]{0pt}{1.7ex}}

% Group labels
\def\evergrouplabelstrut{\rule{0pt}{1.5ex}}
\def\evergrouplabelsep{0pt}

% Column header
\def\groupmemheaderstrut{\rule[-0.4ex]{0pt}{2.0ex}}

\hfuzz=3pt

\ifdefined\dashlinedash
  \setlength\dashlinedash{0.6pt}
  \setlength\dashlinegap{1pt}
\fi

\begin{tabularx}{\dimexpr\textwidth-3pt\relax}{
  @{}l*{7}{>{\centering\arraybackslash}X}@{}
}
\toprule

\groupmemheaderstrut
\hspace{2.5pt}Method
& Multi-Hop
& Update
& Ambiguity
& Implicit
& Temporal
& Abstention
& Average
\\
\midrule

% ==================== GPT-5.5 ====================
\multicolumn{8}{>{\columncolor{EverGPTBg}}c}{%
  \evermodelbarstrut
  \evergptmodelicon\hspace{0.45em}\textit{GPT-5.5}%
} \\
\midrule

\hspace{2.5pt}FullContext
& -- & -- & -- & -- & -- & -- & -- \\

% -------------------- Structured Memory --------------------
\evergroupspacedash
\multicolumn{8}{l}{
  \evergrouplabelstrut
  \hspace{4.5pt}\textit{Structured Memory}
} \\[\evergrouplabelsep]

\hspace{7.5pt}Mem0
& 39.56 & 24.30 & 29.25 & 40.82
& 51.23 & 77.70 & 45.64 \\

\hspace{7.5pt}MemOS
& 20.88 & 14.02 & 26.42 & 28.57
& 20.37 & 74.82 & 31.14 \\

% -------------------- Self-Evolving Memory --------------------
\evergroupspacedash
\multicolumn{8}{l}{
  \evergrouplabelstrut
  \hspace{4.5pt}\textit{Self-Evolving Memory}
} \\[\evergrouplabelsep]

\hspace{7.5pt}A-MEM
& 39.01 & 31.78 & 32.08 & 63.27
& 30.25 & 69.78 & 42.42 \\

\hspace{7.5pt}MemRL
& 29.12 & 30.84 & 31.13 & 61.22
& 19.14 & \textbf{84.17} & 39.87 \\

% -------------------- Human-Designed Harness --------------------
\evergroupspacedash
\multicolumn{8}{l}{
  \evergrouplabelstrut
  \hspace{4.5pt}\textit{Human-Designed Harness}
} \\[\evergrouplabelsep]

\hspace{7.5pt}Codex
& 60.43 & 31.77 & 47.16 & 48.97
& 62.96 & 66.18 & 55.43 \\

\hspace{7.5pt}Claude Code
& 43.96 & 28.30 & 36.89 & 42.86
& 44.44 & 63.31 & 44.95 \\

% -------------------- EvoGroup --------------------
\evergroupspacedash
\multicolumn{8}{l}{
  \evergrouplabelstrut
  \hspace{4.5pt}\textit{EvoGroup}
} \\[\evergrouplabelsep]

\hspace{7.5pt}$H_0$
& 57.69
& \textbf{70.09}
& \textbf{50.00}
& 77.55
& 72.22
& 52.52
& 61.88 \\

\hspace{7.5pt}$H_{*}$
& \textbf{64.84}
& 67.29
& 42.45
& \textbf{79.59}
& \textbf{77.78}
& 54.68
& \textbf{63.89} \\

\midrule

% ==================== DeepSeek-V4-Flash ====================
\multicolumn{8}{>{\columncolor{EverModelBlue}}c}{%
  \evermodelbarstrut
  \everdeepseekmodelicon
  \hspace{0.4em}
  \textit{\textbf{DeepSeek-V4-Flash}}%
} \\
\midrule

\hspace{2.5pt}FullContext
& -- & -- & -- & -- & -- & -- & -- \\

% -------------------- Structured Memory --------------------
\evergroupspacedash
\multicolumn{8}{l}{
  \evergrouplabelstrut
  \hspace{4.5pt}\textit{Structured Memory}
} \\[\evergrouplabelsep]

\hspace{7.5pt}Mem0
& 34.07 & 14.02 & 28.30 & 42.86
& 38.27 & 92.09 & 42.68 \\

\hspace{7.5pt}MemOS
& 17.03 & 8.41 & 18.87 & 22.45
& 17.90 & 91.37 & 30.47 \\

% -------------------- Self-Evolving Memory --------------------
\evergroupspacedash
\multicolumn{8}{l}{
  \evergrouplabelstrut
  \hspace{4.5pt}\textit{Self-Evolving Memory}
} \\[\evergrouplabelsep]

\hspace{7.5pt}A-MEM
& 34.38 & 26.53 & 25.00 & 64.71
& 25.61 & 80.95 & 40.69 \\

\hspace{7.5pt}MemRL
& 25.27 & 21.50 & 18.87 & 53.06
& 16.05 & \textbf{94.96} & 36.64 \\

% -------------------- Human-Designed Harness --------------------
\evergroupspacedash
\multicolumn{8}{l}{
  \evergrouplabelstrut
  \hspace{4.5pt}\textit{Human-Designed Harness}
} \\[\evergrouplabelsep]

\hspace{7.5pt}Codex
& 75.27
& \textbf{85.05}
& 54.72
& \textbf{81.63}
& 70.99
& 92.81
& \textbf{76.51} \\

\hspace{7.5pt}Claude Code
& 68.68 & 57.01 & 46.23 & 61.22
& 67.90 & 89.21 & 66.98 \\

% -------------------- EvoGroup --------------------
\evergroupspacedash
\multicolumn{8}{l}{
  \evergrouplabelstrut
  \hspace{4.5pt}\textit{EvoGroup}
} \\[\evergrouplabelsep]

\hspace{7.5pt}$H_0$
& \textbf{82.42}
& 70.09
& 54.72
& 65.31
& \textbf{77.78}
& 79.86
& 74.09 \\

\hspace{7.5pt}$H_{*}$
& 74.73
& 65.42
& \textbf{62.26}
& 73.47
& 74.69
& 72.66
& 71.14 \\

\bottomrule
\end{tabularx}
\end{table*}

\begin{table*}[t]
\centering
\small
\caption{\textbf{Effectiveness of bucketed analysis} in producing useful evolution signals. \textbf{Correct.} measures the consistency between the analyzed findings and the actual task issues; \textbf{Adopted} indicates that the Evolve Agent accepts a finding and uses it to update the harness; \textbf{Denied} indicates that the Evolve Agent considers the finding incorrect and explicitly rejects it; and \textbf{Unused} indicates that the finding is not selected for a harness update. \textbf{Repair} measures the proportion of affected tasks successfully repaired by adopted findings. All the metrics are reported on a 0--100 scale.
}
\setlength{\tabcolsep}{1.6pt}
\renewcommand{\arraystretch}{1.4}

\begin{minipage}[t]{0.495\textwidth}
\vspace{0pt}
\centering
{\scriptsize (a) DeepSeek-V4-Flash}\par\vspace{0pt}

\resizebox{0.95\linewidth}{!}{%
\begin{tabular}{lccccc}
\toprule
\multirow{2}{*}{\textbf{Iter.}} &
\multirow{2}{*}{\textbf{Correct.}} &
\multicolumn{3}{c}{\textbf{Adoption}} &
\multirow{2}{*}{\textbf{Repair}} \\
\cmidrule(lr){3-5}
& & \textit{Adopted} & \textit{Denied} & \textit{Unused} & \\
\midrule

\EvoIterationKey{EvoBlue}{Iter. 1} &
\EvoMetricBar{EvoBlue}{0.875}{87.50} &
\EvoMetricBar{EvoBlue}{0.625}{62.50} &
\EvoMetricBar{EvoBlue}{0.000}{0.00} &
\EvoMetricBar{EvoBlue}{0.375}{37.50} &
\EvoMetricBar{EvoBlue}{0.594}{59.40} \\

\EvoIterationKey{EvoGreen}{Iter. 2} &
\EvoMetricBar{EvoGreen}{0.889}{88.90} &
\EvoMetricBar{EvoGreen}{0.667}{66.70} &
\EvoMetricBar{EvoGreen}{0.000}{0.00} &
\EvoMetricBar{EvoGreen}{0.333}{33.30} &
\EvoMetricBar{EvoGreen}{0.571}{57.10} \\

\EvoIterationKey{EvoRed}{Iter. 3} &
\EvoMetricBar{EvoRed}{1.000}{100.00} &
\EvoMetricBar{EvoRed}{0.571}{57.10} &
\EvoMetricBar{EvoRed}{0.143}{14.30} &
\EvoMetricBar{EvoRed}{0.286}{28.60} &
\EvoMetricBar{EvoRed}{0.414}{41.40} \\

\EvoIterationKey{EvoOrange}{Iter. 4} &
\EvoMetricBar{EvoOrange}{1.000}{100.00} &
\EvoMetricBar{EvoOrange}{0.667}{66.70} &
\EvoMetricBar{EvoOrange}{0.000}{0.00} &
\EvoMetricBar{EvoOrange}{0.333}{33.30} &
\EvoMetricBar{EvoOrange}{0.545}{54.50} \\

\EvoIterationKey{EvoPurple}{Iter. 5} &
\EvoMetricBar{EvoPurple}{0.714}{71.40} &
\EvoMetricBar{EvoPurple}{0.571}{57.10} &
\EvoMetricBar{EvoPurple}{0.143}{14.30} &
\EvoMetricBar{EvoPurple}{0.286}{28.60} &
\EvoMetricBar{EvoPurple}{0.576}{57.60} \\

\bottomrule
\end{tabular}%
}
\end{minipage}\hfill
\begin{minipage}[t]{0.495\textwidth}
\vspace{0pt}
\centering
{\scriptsize (b) GPT-5.5}\par\vspace{0pt}

\resizebox{0.95\linewidth}{!}{%
\begin{tabular}{lccccc}
\toprule
\multirow{2}{*}{\textbf{Iteration.}} &
\multirow{2}{*}{\textbf{Correct.}} &
\multicolumn{3}{c}{\textbf{Adoption}} &
\multirow{2}{*}{\textbf{Repair}} \\
\cmidrule(lr){3-5}
& & \textit{Adopted} & \textit{Denied} & \textit{Unused} & \\
\midrule

\EvoIterationKey{EvoBlue}{Iter. 1} &
\EvoMetricBar{EvoBlue}{1.000}{100.00} &
\EvoMetricBar{EvoBlue}{0.875}{87.50} &
\EvoMetricBar{EvoBlue}{0.000}{0.00} &
\EvoMetricBar{EvoBlue}{0.125}{12.50} &
\EvoMetricBar{EvoBlue}{0.400}{40.00} \\

\EvoIterationKey{EvoGreen}{Iter. 2} &
\EvoMetricBar{EvoGreen}{1.000}{100.00} &
\EvoMetricBar{EvoGreen}{0.625}{62.50} &
\EvoMetricBar{EvoGreen}{0.000}{0.00} &
\EvoMetricBar{EvoGreen}{0.375}{37.50} &
\EvoMetricBar{EvoGreen}{0.452}{45.20} \\

\EvoIterationKey{EvoRed}{Iter. 3} &
\EvoMetricBar{EvoRed}{1.000}{100.00} &
\EvoMetricBar{EvoRed}{0.857}{85.70} &
\EvoMetricBar{EvoRed}{0.000}{0.00} &
\EvoMetricBar{EvoRed}{0.143}{14.30} &
\EvoMetricBar{EvoRed}{0.686}{68.60} \\

\EvoIterationKey{EvoOrange}{Iter. 4} &
\EvoMetricBar{EvoOrange}{1.000}{100.00} &
\EvoMetricBar{EvoOrange}{0.625}{62.50} &
\EvoMetricBar{EvoOrange}{0.000}{0.00} &
\EvoMetricBar{EvoOrange}{0.375}{37.50} &
\EvoMetricBar{EvoOrange}{0.444}{44.40} \\

\EvoIterationKey{EvoPurple}{Iter. 5} &
\EvoMetricBar{EvoPurple}{1.000}{100.00} &
\EvoMetricBar{EvoPurple}{0.500}{50.00} &
\EvoMetricBar{EvoPurple}{0.000}{0.00} &
\EvoMetricBar{EvoPurple}{0.500}{50.00} &
\EvoMetricBar{EvoPurple}{0.333}{33.30} \\

\bottomrule
\end{tabular}%
}
\end{minipage}
\label{tab:adoption-repair}
\end{table*}

\begin{figure}[htpb]
\centering
\includegraphics[width=\linewidth]{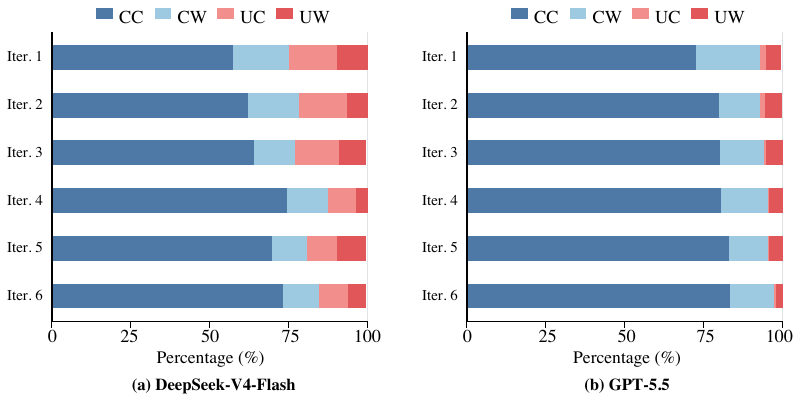}
\vspace{-1em}
\caption{\textbf{Distribution of trajectory types across self-evolution for DeepSeek-V4-Flash and GPT-5.5.} \textbf{CC} denotes confident-and-correct trajectories, representing routine successful behavior. \textbf{CW} denotes confident-but-wrong trajectories, revealing potentially undetected reasoning failures or systematic answer bias. \textbf{UC} denotes unconfident-but-correct trajectories, capturing fragile successes that the current harness may not sufficiently support. \textbf{UW} denotes unconfident-and-wrong trajectories, capturing failures that the agent can partially recognize.}
\label{fig:Confidence}
\end{figure}

\subsection{Efficiency Analysis}
\label{sec:efficiency-analysis}

\textbf{Metric definitions.}
We separate online per-question cost from one-time preparation. Online metrics are \textit{Tokens per Question} (Tok.), \textit{Tool Calls per Question} (Calls), and \textit{Question Runtime} (QA), each averaged over completed trials, including both correct and incorrect answers. Trials that terminate because of a system-level failure, timeout, crash, exception, or missing output are excluded from all three averages. Tok. is the sum of uncached input tokens, cached input tokens, and output tokens attributed to a question across all model calls. Each independently issued executable tool request counts as one call, and parallel requests are counted individually. Requests made by sub-agents are attributed to the parent question. Creating a sub-agent does not itself count as a tool request. QA time is the elapsed wall-clock time from the start of the question until completion and termination. \textit{Cold-Start Time} (Cold) is a separate one-time preparation cost before answering. A dash denotes an unavailable or inapplicable measurement, not zero cost.

\noindent\textbf{Cost profiles.}
Tables~\ref{tab:evermembench-efficiency} and~\ref{tab:groupmembench-efficiency}
show that EvoGroup keeps its per-question token cost within a relatively narrow range of
111K--200K, whereas Codex and Claude Code range from 66K to 668K.
$H_*$ uses fewer tokens than Codex in all four settings, with reductions ranging from
0.8\% to 81.6\%, and uses fewer tokens than Claude Code in three of the four settings.
The only exception is EverMemBench with DeepSeek-V4-Flash, where Claude Code uses 66K
tokens per question compared with 123K for $H_*$.
Evolution also reduces the number of tool calls on EverMemBench from 14.28 to 12.87 with
GPT-5.5 and from 16.47 to 12.57 with DeepSeek-V4-Flash.
With DeepSeek-V4-Flash, $H_{*}$ makes fewer tool calls than either human-designed harness
on both benchmarks, including 13.45 calls compared with Codex's 27.59 on GroupMemBench.
Overall, EvoGroup maintains relatively stable per-question token and tool-call costs while
avoiding the large cost variation observed across the human-designed harnesses.

\noindent\textbf{Runtime profiles.}
Tables~\ref{tab:evermembench-efficiency} and~\ref{tab:groupmembench-efficiency}
report the per-question answering time and cold-start overhead in seconds.
Compared with Codex and Claude Code, $H_*$ reduces question runtime by $60.1\%$--$83.3\%$ and $77.4\%$--$85.9\%$, respectively.
On EverMemBench, evolution further reduces the answering time from
$23.36$~s to $18.44$~s with GPT-5.5 and from $16.67$~s to $14.82$~s
with DeepSeek-V4-Flash.
Memory methods answer each question in $2.13$--$8.69$~s, but their
cold-start time ranges from $1{,}241$~s to $158{,}456$~s.
In contrast, the cold-start time of  $H_*$ is only
$0.074$--$0.181$~s.
Evolution reduces the cold-start time relative to $H_0$ in all four
settings, including a reduction from $0.589$~s to $0.181$~s on
EverMemBench with GPT-5.5.
Overall, EvoGroup answers questions several times faster than the
human-designed harnesses while keeping its cold-start time below one second.

\begin{table*}[t]
\centering

% ==================== Model colors ====================
\definecolor{EverGPTBg}{HTML}{E5EFE9}
\definecolor{EverDeepSeekBg}{HTML}{E6EDFC}

% ==================== Model logos ====================
\def\evergptmodelicon{%
  \raisebox{-0.15ex}{%
    \includegraphics[height=1.90ex]{figures/logos/openai.pdf}%
  }%
}
\def\everdeepseekmodelicon{%
  \raisebox{-0.15ex}{%
    \includegraphics[height=1.90ex]{figures/logos/deepseek.pdf}%
  }%
}
\def\effmodelbarstrut{\rule[-0.8ex]{0pt}{3.2ex}}

% ==================== Dashed group separator ====================
\def\effgroupsep{%
  \addlinespace[2pt]
  \arrayrulecolor{black!35}
  \hdashline[0.8pt/1.5pt]
  \arrayrulecolor{black}
  \addlinespace[1pt]
}

% ==================== Left: EverMemBench ====================
\begin{minipage}[t]{0.485\textwidth}
\vspace{0pt}
\centering

\captionof{table}{\textbf{Efficiency on EverMemBench}.
Tok. (K) denotes the average number of tokens consumed per question
(in thousands), Calls denotes the average number of tool calls per
question, QA denotes the average time required to complete each question,
and Cold denotes the one-time preparation cost before answering, i.e., the time required
for the initial Harness execution. QA and Cold are measured in seconds.}
\label{tab:evermembench-efficiency}

\vspace{2pt}
\scriptsize
\renewcommand{\arraystretch}{1.08}

\begin{tabularx}{\linewidth}{
    @{}
    >{\raggedright\arraybackslash}X
    >{\raggedleft\arraybackslash}p{0.105\linewidth}
    >{\raggedleft\arraybackslash}p{0.105\linewidth}
    >{\raggedleft\arraybackslash}p{0.105\linewidth}
    >{\raggedleft\arraybackslash}p{0.180\linewidth}
    @{}
}
\toprule
\textbf{Method}
& \multicolumn{2}{c}{\textbf{Cost / Question}}
& \multicolumn{2}{c}{\textbf{Runtime}} \\
\cmidrule(lr){2-3}
\cmidrule(lr){4-5}
& \textbf{Tok.}
& \textbf{Calls}
& \textbf{QA}
& \textbf{Cold} \\
\midrule

% -------------------- GPT-5.5 --------------------
\multicolumn{5}{>{\columncolor{EverGPTBg}}c}{%
  \effmodelbarstrut
  \evergptmodelicon\hspace{0.45em}\textbf{\textit{GPT-5.5}}%
} \\

\addlinespace[1.5pt]
\multicolumn{5}{l}{\textit{Structured Memory}} \\[1pt]
\hspace{0.5em}Mem0
& 1.096 & -- & 8.14 & 54,884.28 \\
\hspace{0.5em}MemOS
& 1.315 & -- & 8.31 & 3,162.48 \\

\effgroupsep
\multicolumn{5}{l}{\textit{Self-Evolving Memory}} \\[1pt]
\hspace{0.5em}A-MEM
& 11.680 & -- & 2.34 & 158,123.86 \\
\hspace{0.5em}MemRL
& 3.006 & -- & 4.34 & 8,620.30 \\

\effgroupsep
\multicolumn{5}{l}{\textit{Human-Designed Harness}} \\[1pt]
\hspace{0.5em}Codex
& 216 & 8.56 & 46.23 & 0.051 \\
\hspace{0.5em}Claude Code
& 205 & 7.88 & 81.75 & 0.034 \\

\effgroupsep
\multicolumn{5}{l}{\textit{EvoGroup}} \\[1pt]
\hspace{0.5em}$H_0$
& 145 & 14.28 & 23.36 & 0.589 \\
\hspace{0.5em}$H_{*}$
& 157 & 12.87 & 18.44 & 0.181 \\

\midrule

% -------------------- DeepSeek-V4-Flash --------------------
\multicolumn{5}{>{\columncolor{EverDeepSeekBg}}c}{%
  \effmodelbarstrut
  \everdeepseekmodelicon\hspace{0.45em}%
  \textbf{\textit{DeepSeek-V4-Flash}}%
} \\

\addlinespace[1.5pt]
\multicolumn{5}{l}{\textit{Structured Memory}} \\[1pt]
\hspace{0.5em}Mem0
& 1.105 & -- & 2.13 & 54,456.28 \\
\hspace{0.5em}MemOS
& 1.318 & -- & 3.30 & 3,147.12 \\

\effgroupsep
\multicolumn{5}{l}{\textit{Self-Evolving Memory}} \\[1pt]
\hspace{0.5em}A-MEM
& 10.909 & -- & 2.21 & 158,456.23 \\
\hspace{0.5em}MemRL
& 3.075 & -- & 3.53 & 8,665.31 \\

\effgroupsep
\multicolumn{5}{l}{\textit{Human-Designed Harness}} \\[1pt]
\hspace{0.5em}Codex
& 668 & 16.96 & 56.64 & 0.021 \\
\hspace{0.5em}Claude Code
& 66 & 17.32 & 74.56 & 0.014 \\

\effgroupsep
\multicolumn{5}{l}{\textit{EvoGroup}} \\[1pt]
\hspace{0.5em}$H_0$
& 111 & 16.47 & 16.67 & 0.107\\
\hspace{0.5em}$H_{*}$
& 123 & 12.57 & 14.82 & 0.074 \\

\bottomrule
\end{tabularx}
\end{minipage}
\hfill
% ==================== Right: GroupMemBench ====================
\begin{minipage}[t]{0.485\textwidth}
\vspace{0pt}
\centering

\captionof{table}{\textbf{Efficiency on GroupMemBench}.
Tok. (K) denotes the average number of tokens consumed per question
(in thousands), Calls denotes the average number of tool calls per
question, QA denotes the average time required to complete each question,
and Cold denotes the one-time preparation cost before answering, i.e., the time required
for the initial Harness execution. QA and Cold are measured in seconds.}
\label{tab:groupmembench-efficiency}

\vspace{2pt}
\scriptsize
\renewcommand{\arraystretch}{1.08}

\begin{tabularx}{\linewidth}{
    @{}
    >{\raggedright\arraybackslash}X
    >{\raggedleft\arraybackslash}p{0.105\linewidth}
    >{\raggedleft\arraybackslash}p{0.105\linewidth}
    >{\raggedleft\arraybackslash}p{0.105\linewidth}
    >{\raggedleft\arraybackslash}p{0.180\linewidth}
    @{}
}
\toprule
\textbf{Method}
& \multicolumn{2}{c}{\textbf{Cost / Question}}
& \multicolumn{2}{c}{\textbf{Runtime}} \\
\cmidrule(lr){2-3}
\cmidrule(lr){4-5}
& \textbf{Tok.}
& \textbf{Calls}
& \textbf{QA}
& \textbf{Cold} \\
\midrule

% -------------------- GPT-5.5 --------------------
\multicolumn{5}{>{\columncolor{EverGPTBg}}c}{%
  \effmodelbarstrut
  \evergptmodelicon\hspace{0.45em}\textbf{\textit{GPT-5.5}}%
} \\

\addlinespace[1.5pt]
\multicolumn{5}{l}{\textit{Structured Memory}} \\[1pt]
\hspace{0.5em}Mem0
& 0.984 & -- & 6.63 & 1,241 \\
\hspace{0.5em}MemOS
& 0.985 & -- & 4.58 & 3,930 \\

\effgroupsep
\multicolumn{5}{l}{\textit{Self-Evolving Memory}} \\[1pt]
\hspace{0.5em}A-MEM
& 14.611 & -- & 4.69 & 145,654 \\
\hspace{0.5em}MemRL
& 3.878 & -- & 8.69 & 24,920 \\

\effgroupsep
\multicolumn{5}{l}{\textit{Human-Designed Harness}} \\[1pt]
\hspace{0.5em}Codex
& 125 & 4.36 & 93.24 & 0.092 \\
\hspace{0.5em}Claude Code
& 219 & 7.34 & 157.31 & 0.064 \\

\effgroupsep
\multicolumn{5}{l}{\textit{EvoGroup}} \\[1pt]
\hspace{0.5em}$H_0$
& 120 & 14.36 & 20.13 & 0.145 \\
\hspace{0.5em}$H_{*}$
& 124 & 14.46 & 22.25 & 0.116 \\

\midrule

% -------------------- DeepSeek-V4-Flash --------------------
\multicolumn{5}{>{\columncolor{EverDeepSeekBg}}c}{%
  \effmodelbarstrut
  \everdeepseekmodelicon\hspace{0.45em}%
  \textbf{\textit{DeepSeek-V4-Flash}}%
} \\

\addlinespace[1.5pt]
\multicolumn{5}{l}{\textit{Structured Memory}} \\[1pt]
\hspace{0.5em}Mem0
& 1.106 & -- & 4.48 & 1,253 \\
\hspace{0.5em}MemOS
& 1.090 & -- & 3.39 & 3,930 \\

\effgroupsep
\multicolumn{5}{l}{\textit{Self-Evolving Memory}} \\[1pt]
\hspace{0.5em}A-MEM
& 4.089 & -- & 4.42 & 131,314 \\
\hspace{0.5em}MemRL
& 16.523 & -- & 5.79 & 24,857 \\

\effgroupsep
\multicolumn{5}{l}{\textit{Human-Designed Harness}} \\[1pt]
\hspace{0.5em}Codex
& 554 & 27.59 & 166.77 & 0.078 \\
\hspace{0.5em}Claude Code
& 477 & 16.54 & 138.34 & 0.035 \\

\effgroupsep
\multicolumn{5}{l}{\textit{EvoGroup}} \\[1pt]
\hspace{0.5em}$H_0$
& 119 & 14.68 & 28.91 & 0.142 \\
\hspace{0.5em}$H_{*}$
& 200 & 13.45 & 27.86 & 0.098 \\

\bottomrule
\end{tabularx}
\end{minipage}

\end{table*}

\section{Experimental Setup}
\label{app:experimental-setup}
This appendix provides the benchmark composition, evaluation procedure, evolution settings, and supplementary results. The protocols below expand the experimental description in Section~\ref{sec:experiments}.

\subsection{Benchmarks and Data Partitioning}
\label{app:benchmarks-and-splits}
\textbf{GroupMemBench.}
GroupMemBench tests memory in workplace conversations with several users and channels. To answer a question, the Agent must find information from earlier conversations and keep track of who said it. The 745 questions cover multi-hop reasoning, knowledge updates, temporal reasoning, user-implicit reasoning, term ambiguity, and abstention. Table~\ref{tab:domain_distribution} gives the number of questions in each domain and question type.

\begin{table}[t]
\centering
\caption{Overview of GroupMemBench and the distribution of question types across domains.}
\label{tab:domain_distribution}

\renewcommand{\arraystretch}{1.35}
\setlength{\tabcolsep}{5.5pt}

\resizebox{\columnwidth}{!}{%
\begin{tabular}{lrrrrrrr}
\toprule
\textbf{Domain}
& \textbf{Multi-Hop}
& \textbf{Know. Update}
& \textbf{Temporal}
& \textbf{User-Implicit}
& \textbf{Term Ambig.}
& \textbf{Abstention}
& \textbf{Total} \\
\midrule

Finance
& 48 & 32 & 45 & 15 & 45 & 29 & 214 \\

Healthcare
& 48 & 17 & 37 & 2 & 7 & 34 & 145 \\

Manufacturing
& 45 & 22 & 43 & 4 & 11 & 48 & 173 \\

Technology
& 41 & 36 & 37 & 28 & 43 & 28 & 213 \\

\midrule
\textbf{Total}
& \textbf{182}
& \textbf{107}
& \textbf{162}
& \textbf{49}
& \textbf{106}
& \textbf{139}
& \textbf{745} \\

\bottomrule
\end{tabular}%
}
\vspace{-1.0em}
\end{table}

\noindent\textbf{EverMemBench.}
EverMemBench tests memory over long conversations among several people. It covers factual recall, memory awareness, and user-profile understanding. Its questions ask the Agent to recall facts, use relevant past information, track changes, and infer user characteristics. Table~\ref{tab:evermembench_distribution} shows the nine question types and their counts. The full benchmark has 2,400 questions.

\begin{table}[t]
\centering
\caption{Overview of EverMemBench and the distribution of question types across its evaluation dimensions.}
\label{tab:evermembench_distribution}

\small
\setlength{\tabcolsep}{6pt}
\renewcommand{\arraystretch}{1.05}

\begin{tabular}{llr}
\toprule
\textbf{Dimension}
& \textbf{Question Type}
& \textbf{Questions} \\
\midrule

\multirow{3}{*}{Fine-Grained Recall}
& Single-Hop (Single) & 213 \\
& Multi-Hop (Multi)   & 249 \\
& Temporal (Temp.)    & 300 \\
\cmidrule(lr){1-3}

\multirow{3}{*}{Memory Awareness}
& Constraint (Const.)   & 402 \\
& Proactivity (Proact.) & 427 \\
& Update                & 268 \\
\cmidrule(lr){1-3}

\multirow{3}{*}{Profile Understanding}
& Style              & 176 \\
& Skill              & 169 \\
& Role               & 196 \\
\midrule

Total
& --
& \textbf{2,400} \\

\bottomrule
\end{tabular}
\vspace{-1.0em}
\end{table}

\textbf{Evolution and evaluation partitions.}
To ensure a rigorous evaluation of EvoGroup's effectiveness, we explicitly separate the data used for evolution from the data used for held-out evaluation. We randomly select 720 questions, or 30\% of EverMemBench, as the held-in training set for evolution. We sample from each of the nine question types so that their shares remain close to those in the full dataset. The 720 held-in questions remain fixed across all evolution rounds.  Every held-in question is attempted exactly once and scored under pass@1. The remaining 1,680 questions form a separate held-out validation set, which is likewise evaluated under pass@1 with one attempt per question. Validation questions are never used to suggest harness changes or select checkpoints. For transfer, we apply the harness evolved on the 720-question held-in training set of EverMemBench to GroupMemBench without further evolution or tuning.

\subsection{Evaluation Metrics and Judging}
\label{app:evaluation-metrics}
All experimental results are strictly evaluated using the official evaluation scripts provided by each benchmark, ensuring that our evaluation protocol remains consistent with the original benchmarks. Based on these evaluations, we report both \textbf{question-level accuracy} and \textbf{category-level accuracy}. Let $\mathcal{D}$ be an evaluation set containing $N$ questions, and let $y_i \in \{0,1\}$ denote the correctness indicator for question $i$. We compute
\begin{equation}
\mathrm{Acc}(\mathcal{D})=\frac{100}{N}\sum_{i\in\mathcal{D}}y_i,
\qquad
\mathrm{Acc}_k=\frac{100}{N_k}\sum_{i\in\mathcal{D}_k}y_i,
\label{eq:appendix-accuracy}
\end{equation}
where $\mathcal{D}_k$ contains $N_k$ questions of type $k$. Overall accuracy is computed over individual questions, equivalently as
\begin{equation}
\mathrm{Acc}(\mathcal{D})
=
\frac{\sum_k N_k\mathrm{Acc}_k}{\sum_k N_k}.
\end{equation}
Therefore, overall accuracy is weighted by the number of questions in each category, rather than computed as the arithmetic mean of category-level accuracies.

For \textbf{EverMemBench}, we follow its official evaluation script: multiple-choice questions are scored by matching the predicted option against the reference option, while open-ended questions are evaluated using a binary semantic-equivalence judgment from a unified \textbf{GPT-5.6-luna judge}. Let $q_i$, $a_i$, and $\hat{a}_i$ denote the question, reference answer, and predicted answer for question $i$, respectively. The correctness indicator is defined as
\begin{equation}
y_i=
\begin{cases}
\mathbb{I}[\hat{a}_i=a_i], & i\in\mathcal{D}_{\mathrm{MC}},\\
J(q_i,a_i,\hat{a}_i), & i\in\mathcal{D}_{\mathrm{OE}}.
\end{cases}
\label{eq:appendix-evermem-scoring}
\end{equation}
Here, $J$ denotes the GPT-5.6-luna judge, where an output of $1$ indicates that the predicted answer is semantically correct and an output of $0$ indicates an incorrect answer.

For \textbf{GroupMemBench}, we likewise follow its official evaluation script and use the GPT-5.6-luna judge uniformly across all six question categories for semantic evaluation, i.e., $y_i=J(q_i,a_i,\hat{a}_i)$. A judgment of $1$ indicates a semantically correct answer, while $0$ indicates an incorrect answer.

\subsection{Model and Runtime Configuration}
\label{app:runtime-settings}
This section summarizes the model configurations, runtime constraints, and evolution settings used in our experiments; detailed hyperparameters and protocol settings are provided in Table~\ref{tab:hyperparameters}.

\begin{table*}[t]
\centering
\caption{Hyperparameter settings for EvoGroup on EverMemBench. We report the DeepSeek-V4-Flash configuration as a representative example. Model-specific runs replace the backing model while retaining the stated protocol limits. All reported runs use GPT-5.6-luna as the shared LLM judge.}
\label{tab:hyperparameters}

\small
\setlength{\tabcolsep}{10pt}
\renewcommand{\arraystretch}{1.15}

\begin{tabular*}{\textwidth}{@{\extracolsep{\fill}}lll}
\toprule
\textbf{Block} & \textbf{Hyperparameter} & \textbf{Value} \\
\midrule

\multirow{10}{*}{Group Agent}
& Backing model & DeepSeek-V4-Flash \\
& Temperature & Provider default \\
& Max context & 200,000 tokens \\
& Max output per call & 8,192 tokens \\
& Max steps & 60 \\
& Max tool calls & 20 \\
& Per-question timeout & 500 s \\
& \texttt{model\_reasoning\_effort} & \texttt{none} \\
& Available tools & \texttt{list\_files}, \texttt{read\_file}, \texttt{grep\_search} \\
& & \texttt{write\_file}, \texttt{create\_file}, \texttt{query\_records} \\
\midrule

\multirow{7}{*}{Analysis Agent}
& Backing model & DeepSeek-V4-Flash \\
& Temperature & 0.4 \\
& \texttt{model\_reasoning\_effort} & \texttt{none} \\
& Concurrent tasks & 8 \\
& Per-question timeout & 900 s \\
& Retry attempts & 3 \\
& Retry backoff & 10 s, doubled after each retry \\
\midrule

\multirow{7}{*}{Evolve Agent}
& Backing model & DeepSeek-V4-Flash \\
& Temperature & 0.7 \\
& \texttt{model\_reasoning\_effort} & \texttt{none} \\
& Max context & 200,000 tokens \\
& Max generation per turn & 32,000 tokens \\
& Max model turns & 500 \\
& Context compaction threshold & 0.75 \\
\midrule

\multirow{4}{*}{LLM-as-a-Judge}
& Backing model & GPT-5.6-luna \\
& Temperature & 0.0 \\
& Retry attempts & 3 \\
& Concurrency & 4 \\
\bottomrule
\end{tabular*}
\end{table*}

\section{Evolution Protocol}
\label{app:evolution-protocol}
\subsection{Trajectory Analysis and Harness Revision}
Let $\mathcal T=\{x_i\}_{i=1}^{N}$ be the fixed evolution questions and $H_t$ be the harness at round $t$. An interaction produces trajectory $\tau_i^t=\operatorname{Interact}(H_t,x_i)$, the Agent's confidence $p_i^t\in[0,1]$, and the judge's answer-correctness label $y_i^t\in\{0,1\}$. The trajectory records the messages, tool requests and observations, memory operations, and final answer. A timeout or infrastructure exception is recorded separately from a judged incorrect answer.

At confidence threshold $\gamma=0.5$, semantic trajectories are assigned to confident-correct (CC), confident-wrong (CW), unconfident-correct (UC), or unconfident-wrong (UW), where ``confident'' means $c_i>\gamma$. Low-confidence trajectories also receive a self-reflection, used as diagnostic evidence without changing their official correctness label. The Analysis Agent first inspects addressable summaries and retrieves full trace segments when it needs to verify a finding. It synthesizes recurring patterns across questions before the Evolve Agent proposes changes to the harness components: skills, prompts, middleware, tools, and memory. Each proposal includes a change manifest and its predicted repair and regression risks.

The result measured at round $t$ belongs to $H_t$. A candidate $H_{t+1}$ is only a proposal until it has been executed and evaluated in a subsequent round. The evolution set is used to select a checkpoint; results on a separate test set are reported independently.

\subsection{Candidate Validation and Capability Transitions}
\label{app:validation-and-transitions}
Each candidate must include a change manifest. Structural checks require parseable files, configuration, prompts, and tool interfaces. An edit must be linked to a verified trajectory or cross-trace finding and must remain task-general: it cannot encode a task ID, a corpus phrase, a reference answer, or a question-specific rule. The task set, judge, model configuration, trace collector, and experimental infrastructure are held fixed. If no candidate passes, $H_{t+1}=H_t$. Among candidates that pass, the stated selection priority is evidence coverage, predicted repair benefit, predicted regression risk, edit scope, and finally manifest completeness. These priorities govern which candidate is tried next; they are not evidence of its empirical effect.

For two consecutively evaluated harnesses, a question can be fail-to-pass (F2P), pass-to-fail (P2F), always-pass (AP), or always-fail (AF). Timeouts and infrastructure exceptions are tracked separately. If the paired question outcomes are available, the corresponding gain, regression, and retention rates are

\begin{equation}
\begin{aligned}
\mathrm{GainRate}_t
&=\frac{N_{\mathrm{F2P}}}{N_{\mathrm{F2P}}+N_{\mathrm{AF}}},\\
\mathrm{RegressionRate}_t
&=\frac{N_{\mathrm{P2F}}}{N_{\mathrm{AP}}+N_{\mathrm{P2F}}},\\
\mathrm{Retention}_t
&=1-\mathrm{RegressionRate}_t.
\end{aligned}
\label{eq:appendix-transition-rates}
\end{equation}

These quantities are defined only where their denominators are nonzero and are computed from paired pass@1 outcomes for the same questions. They supplement, rather than replace, official question-level accuracy.

\end{document}